\documentclass{article}

\usepackage{amsmath, amssymb}
\usepackage{multirow}
\usepackage{threeparttable}
\usepackage[table]{xcolor}
\usepackage{subcaption}

\usepackage[final,eandd]{neurips_2026}

\usepackage{tabularx}
\usepackage[utf8]{inputenc}
\usepackage[T1]{fontenc}
\usepackage{hyperref}
\usepackage{url}
\usepackage{booktabs}
\usepackage{amsfonts}
\usepackage{nicefrac}
\usepackage{microtype}
\usepackage{xcolor}
\usepackage[many]{tcolorbox}
\usepackage{graphicx}
\usepackage{array}
\usepackage{makecell}
\usepackage{enumitem}
\usepackage{ragged2e}

\definecolor{PromptInstColor}{RGB}{0, 90, 160}
\definecolor{PromptMemColor}{RGB}{210, 80, 0}
\definecolor{PromptShortColor}{RGB}{20, 120, 70}

\newtcolorbox{promptbox}[1][]{
    breakable,
    colback=gray!5!white,
    colframe=gray!60!black,
    title=\textbf{Prompt Example},
    fonttitle=\sffamily\bfseries,
    boxrule=0.8pt,
    arc=4pt,
    left=8pt, right=8pt, top=8pt, bottom=8pt,
    #1
}

\title{\textbf{SatNav}: A Scalable Benchmark for Long-Horizon UAV Vision-Language Navigation from Satellite Imagery}

\author{%
  \textbf{Jiajun Jiang}$^{1}$\thanks{Equal contribution. \quad $^{\dagger}$Corresponding author.} \quad \textbf{Chunliang Hua}$^{1}$\footnotemark[1] \quad \textbf{Zichun Chen}$^{2}$ \quad \textbf{Yanxing Wu}$^{2}$ \\
  \textbf{Zeyuan Yang}$^{2}$ \quad \textbf{Jie Song}$^{1,3}$ \quad \textbf{Xiao Hu}$^{1,2,\dagger}$ \\
  \normalfont $^{1}$\,The Hong Kong University of Science and Technology (Guangzhou) \\
  \normalfont $^{2}$\,Low Altitude Space Economy Research Center, \\
  \normalfont International Digital Economy Academy (IDEA) \\
  \normalfont $^{3}$\,The Hong Kong University of Science and Technology
}

\begin{document}

\maketitle

\begin{abstract}
 
    Urban uncrewed aerial vehicle (UAV) vision-language navigation (VLN) requires agents to follow instructions across extended urban spaces, inherently demanding long-term memory and geospatial grounding. However, scaling existing benchmarks remains difficult because of their reliance on costly reconstructed 3D assets, limiting geographic diversity and episode scale.
    To address this, we introduce SatNav, a scalable, long-horizon UAV VLN benchmark built from high-resolution satellite imagery.
    SatNav targets city-level navigation missions and uses satellite crops as approximations of UAV nadir views for visual observations. Through an automated cue-to-episode pipeline, SatNav constructs 118K episodes from 59 scenes across 18 cities, with an average trajectory length of 379\,m. 
    To stress-test long-horizon memory and geospatial reasoning, SatNav defines three task families: {Boundary}, {Landmark}, and {Route}, targeting loop progress tracking, landmark-based spatial grounding, and route following with counting cues. Benchmarking classical VLN agents and recent agents based on large vision-language models (LVLMs) on SatNav shows that city-scale navigation remains challenging.
    We further introduce {SwiftVLN}, a modular framework with switchable memory components, and conduct systematic memory-design ablations. 
    Finally, satellite-to-UAV transfer experiments show that satellite-trained navigation models can operate on real-flight UAV observations, showing the practical relevance of SatNav.
    Our project page: \url{https://eku127.github.io/SatNav/}.

\end{abstract}

\section{Introduction}

Vision-language navigation (VLN) is a foundational problem at the intersection of computer vision, natural language processing, and embodied AI, where agents follow natural-language instructions to navigate through environments~\citep{anderson2018vision,qi2020reverie}.
Many existing benchmarks focus on relatively short-range paths, whereas real-world navigation often requires agents to operate over large areas and extended time horizons. This places stronger demands on memory and spatial reasoning, as agents must relate current observations to past decisions and execute instructions consistently over long-range trajectories~\citep{fang2019scene,wei2025streamvln}. While recent work has begun to explore long-horizon VLN in indoor settings~\citep{song2025towards, lin2025vlnverse}, the limited spatial scale and the restricted navigable area of indoor scenes prevent these benchmarks from fully stress-testing agent memory and temporal consistency over long distances.

In contrast, urban uncrewed aerial vehicle (UAV) navigation offers a natural and challenging testbed for long-horizon VLN, as city-level aerial operations inherently require agents to operate over large areas and extended routes~\citep{otto2018optimization}.
Although this setting has motivated several recent UAV VLN benchmarks~\citep{liu2023aerialvln,gao2025openfly,cai2026airnav}, existing efforts remain limited in both scalability and long-horizon evaluation. First, scalability is constrained by the dependence on reconstructed 3D assets. While such assets support realistic simulation~\citep{liu2023aerialvln, lee2024citynavlanguagegoalaerialnavigation}, scaling them across many cities and routes is costly and labor-intensive, which limits geographic diversity and makes large-scale automated episode generation difficult. Second, long-horizon evaluation remains underdeveloped. Existing benchmarks effectively address short-range instruction following and flight control, but their task designs provide limited coverage of long-range UAV navigation that requires agents to maintain and use memory over extended routes.

To address these limitations, we exploit the observation that long-horizon UAV navigation is often governed by large-scale geospatial structures such as road networks and regional layouts, rather than fine-grained 3D geometry~\citep{ryll2020semantic}.
These structural cues are largely preserved in top-down nadir UAV views and can be well approximated by high-resolution satellite image crops~\citep{dai2023vision}.
This approximation provides a scalable path to city-scale benchmark construction, as satellite imagery is widely available and observations can be generated through simple rotate-and-crop operations rather than complex 3D rendering.

Building on this insight, we introduce \textbf{SatNav}, a benchmark for memory-demanding long-horizon VLN in city-scale UAV environments.
SatNav automatically constructs navigation episodes from paired satellite imagery and OpenStreetMap (OSM) annotations, yielding 118K episodes across 18 cities with an average trajectory length of 379\,m.
Beyond scale, SatNav targets the long-range memory and geospatial reasoning required by urban UAV missions such as patrolling, inspection, and logistics~\citep{otto2018optimization}.
We select three complementary task families motivated by common navigation behaviors in these missions: \textbf{Boundary} evaluates tracking progress around an object or region; \textbf{Landmark} evaluates orienting by visual landmarks and relative spatial cues; and \textbf{Route} evaluates following structured routes and tracking successive decision points to determine when to turn or stop.
Each family provides detailed, natural, and concise instruction styles, enabling evaluation under varying levels of linguistic specificity.
Benchmarking classical VLN methods and recent LVLM-based approaches on SatNav shows that city-scale long-horizon navigation remains challenging.
To support future research, we introduce \textbf{SwiftVLN}, a modular framework featuring switchable memory modules, and utilize it to systematically ablate various memory designs.
Finally, we study satellite-to-UAV transfer and show that models trained on large-scale satellite-view data can transfer to real-flight UAV observations, highlighting the practical relevance of SatNav.

\begin{figure*}[t]
    \centering
    \includegraphics[width=\textwidth]{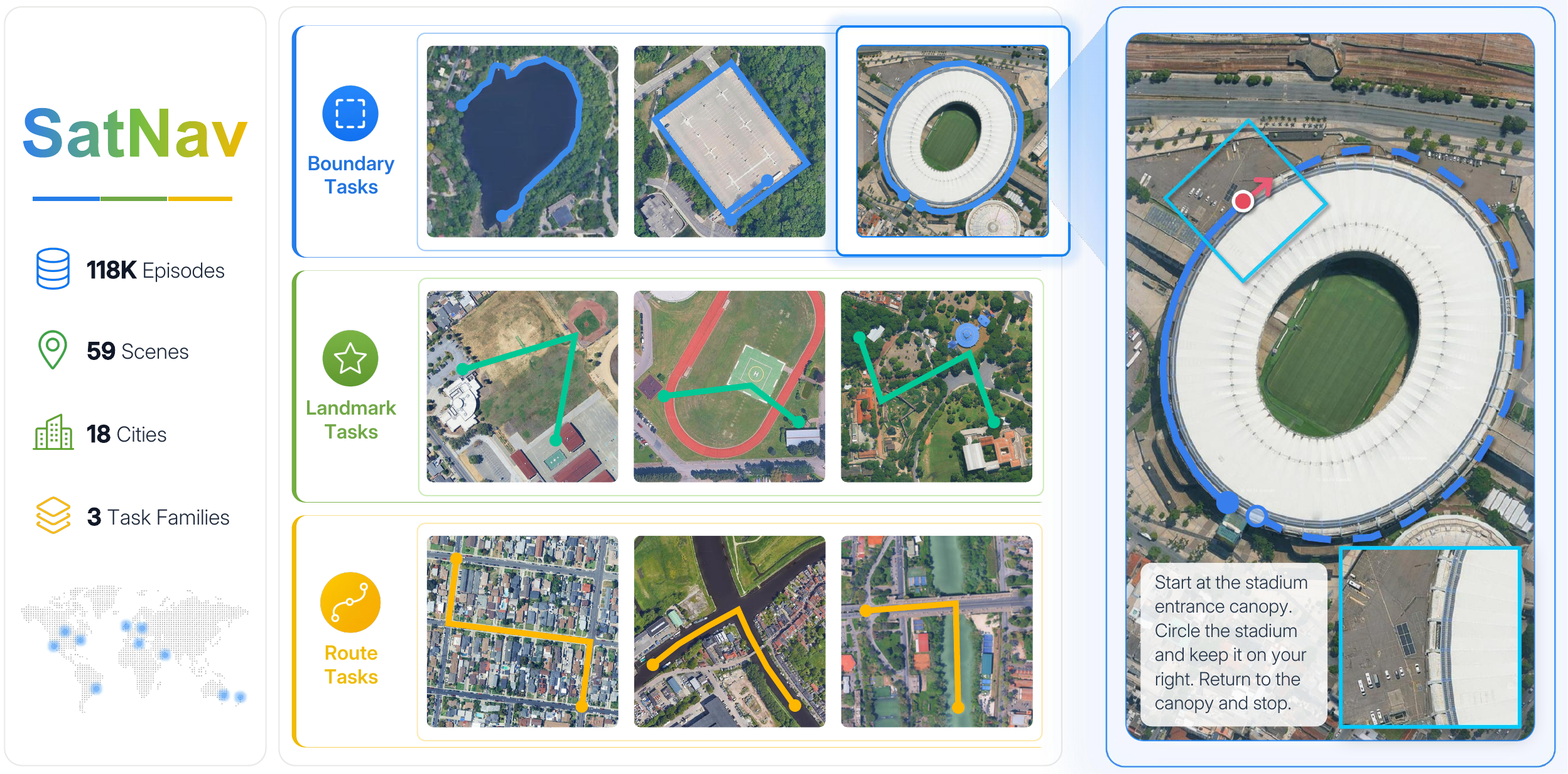}
    \caption{\textbf{Overview of SatNav}. The benchmark spans diverse cities and scenes, and introduces three task families: \textbf{Boundary} (\textit{partial-arc}, \textit{extended-loop}, \textit{full-loop}), \textbf{Landmark} (\textit{one-turn}, \textit{two-turn}), and \textbf{Route} (\textit{road}, \textit{waterway}, \textit{hybrid}). During navigation, the agent receives an instruction and a satellite crop generated by rotate-and-crop operations, then predicts and executes actions in the satellite map.}
    \label{fig:satnav_overview}
    \vspace{0pt}
\end{figure*}

\newpage
We summarize our contributions as follows:
\begin{itemize}[leftmargin=1.5em, labelsep=0.5em]
    \item We introduce \textbf{SatNav}, a scalable benchmark for memory-demanding long-horizon UAV VLN, comprising 118K automatically constructed episodes across 18 cities from satellite imagery and OSM annotations.

    \item We design three complementary task families, namely {Boundary}, {Landmark}, and {Route}, with detailed, natural, and concise instruction styles to evaluate long-term memory and spatial reasoning.

    \item We benchmark classical and LVLM-based VLN models on \textbf{SatNav}, introduce the \textbf{SwiftVLN} framework, study memory-enhancement variants on top of it, and demonstrate successful satellite-to-UAV transfer in real-flight settings.
\end{itemize}

\section{Related Work}

\subsection{Indoor and Aerial VLN Benchmarks}

Indoor VLN benchmarks established the standard instruction-following paradigm and gradually expanded its language, interaction, and control settings~\citep{anderson2018vision,jain2019stay,qi2020reverie,ku2020room,thomason2020vision,krantz2020beyond}. Recent indoor benchmarks further introduce multi-stage and iterative settings to study longer-horizon navigation~\citep{song2025towards,krantz2023iterative,hong2025general}. However, indoor scenes remain limited in spatial extent and topological diversity, making it difficult to stress-test memory and temporal consistency over truly long-range routes.
Extending beyond indoor environments, aerial and UAV VLN benchmarks have advanced language-guided navigation into urban settings~\citep{liu2023aerialvln,gao2025openfly,cai2026airnav}. Other benchmarks further investigate specific high-level missions, including target search~\citep{wang2024towards,xiao2025uav, lee2024citynavlanguagegoalaerialnavigation} and safety-aware navigation~\citep{guo2026huge}. Despite this progress, many existing aerial benchmarks still rely on complex 3D assets or simulation pipelines and often emphasize local instruction following and flight control, leaving the systematic evaluation of long-range memory and geospatial reasoning over urban routes unaddressed.

\subsection{Memory Modeling in Long-Horizon VLN}

In parallel with the scaling of VLN environments, VLN methods have rapidly evolved, particularly with the adoption of LVLMs, which have shown strong generalization across navigation settings~\citep{zheng2024towards, cheng2024navila, zhang2025embodied}. To address long-horizon navigation, recent methods have increasingly emphasized memory modeling to better exploit historical observations. Representative strategies include map-based memory architectures~\citep{zhang2025mapnav}, selective history-frame sampling~\citep{wei2025streamvln, cai2026airnav, zheng2026onfly}, pose-aware history encoding~\citep{gopinathan2024stratxplore}, and token-level history compression~\citep{zhang2024uni, gao2025openfly, jiang2025longfly}. While these methods show promising results, existing evaluations provide limited support for systematically isolating and comparing memory mechanisms under city-scale, long-horizon navigation settings.

\section{SatNav Bench}
\label{sec:satnav_bench}

\subsection{Task Definition}
\label{sec:task_definition}

SatNav formulates satellite-view VLN as an instruction-following sequential decision-making task.
Each episode is specified by a natural-language instruction $\mathcal{I}$.
At each decision step $t$, the agent receives a satellite crop as its current observation $o_t$, generated from the corresponding satellite map through rotate-and-crop operations, as shown in the right panel of Figure~\ref{fig:satnav_overview}.
Conditioned on $\mathcal{I}$, $o_t$, and the navigation history, the agent predicts either a single action $a_t$ or an action sequence $A_t$, which is then executed in the satellite-map environment.
We use four discrete actions: \texttt{forward} moves the agent $10\,\mathrm{m}$ along its current heading, \texttt{left} and \texttt{right} rotate the heading by $15^\circ$, and \texttt{stop} terminates the episode.
The episode ends when the agent outputs \texttt{stop} or reaches the maximum step budget.
The observation generation process is detailed in Appendix~\ref{app:observation_cropping}.

To evaluate complementary aspects of memory-intensive long-horizon navigation, SatNav defines three task families: \textbf{Boundary}, \textbf{Landmark}, and \textbf{Route}, covering looped-route progress tracking, landmark-based spatial grounding, and sequential route reasoning.

\textbf{Boundary} tasks require the agent to follow the perimeter of a lake, stadium, or building complex. Starting from a point on this closed boundary, the agent navigates along it and stops at an instruction-specified goal.
We define three variants: \textit{partial-arc}, where the agent stops before completing a full loop; \textit{full-loop}, where it stops upon returning to the start after one complete traversal; and \textit{extended-loop}, where it passes the start and follows an additional segment before stopping.
These tasks evaluate long-horizon progress tracking and accurate stopping, requiring the agent to distinguish between intermediate visits and the final arrival at a goal location.

\textbf{Landmark} tasks require the agent to follow a route to a specified destination, using visually distinctive landmarks to determine its heading at turns. The agent follows the instructed movements and, at each turning point, rotates until the referenced landmark appears in the specified region of its egocentric view. For example, ``turn left until the stadium appears in the upper-right of the view, then continue forward'' uses the stadium's position to specify the desired heading after the turn. The agent proceeds through the remaining instructions and stops at the destination.
We define \textit{one-turn} and \textit{two-turn} variants with different numbers of landmark-grounded decisions.
These tasks evaluate landmark grounding and relative spatial reasoning.

\textbf{Route} tasks require the agent to navigate along roads, waterways, or combinations of both, following instructions that specify when to turn or stop. The agent must identify the relevant intersections or junctions along the route, using cues such as ``turn left at the first bridge'' or ``stop at the third crossroad.''
Based on the route composition, we categorize tasks into \textit{road-only}, \textit{waterway-only}, and \textit{hybrid} variants, with the latter combining road and waterway segments.
These tasks evaluate route following, junction-level decision-making, and the ability to track and count successive decision points over long trajectories.

\subsection{Geospatial Data Resources}
\label{sec:data_resources}

SatNav is constructed from scalable 2D geospatial resources rather than reconstructed 3D assets.
Specifically, we use high-resolution satellite imagery (e.g., Google Maps~\citep{googlemaps}) as the visual source and align it with OSM annotations~\citep{openstreetmap} for structured geographic information.
Satellite imagery provides real-world top-down visual observations and appearance-based landmark cues, such as red roofs, while OSM provides structured semantic and geometric cues, such as lakes, roads, and waterways.
Together, these resources support the generation of Boundary, Landmark, and Route episodes across diverse urban environments.
SatNav currently covers 18 cities and 59 scenes, and can be extended by selecting new geographic regions and applying the same data-generation pipeline.
Further details on scene selection are provided in Appendix~\ref{app:scene_collection}.

\subsection{Episode Generation}
\label{sec:episode_generation}

\begin{figure*}[t]
    \centering
    \includegraphics[width=\textwidth]{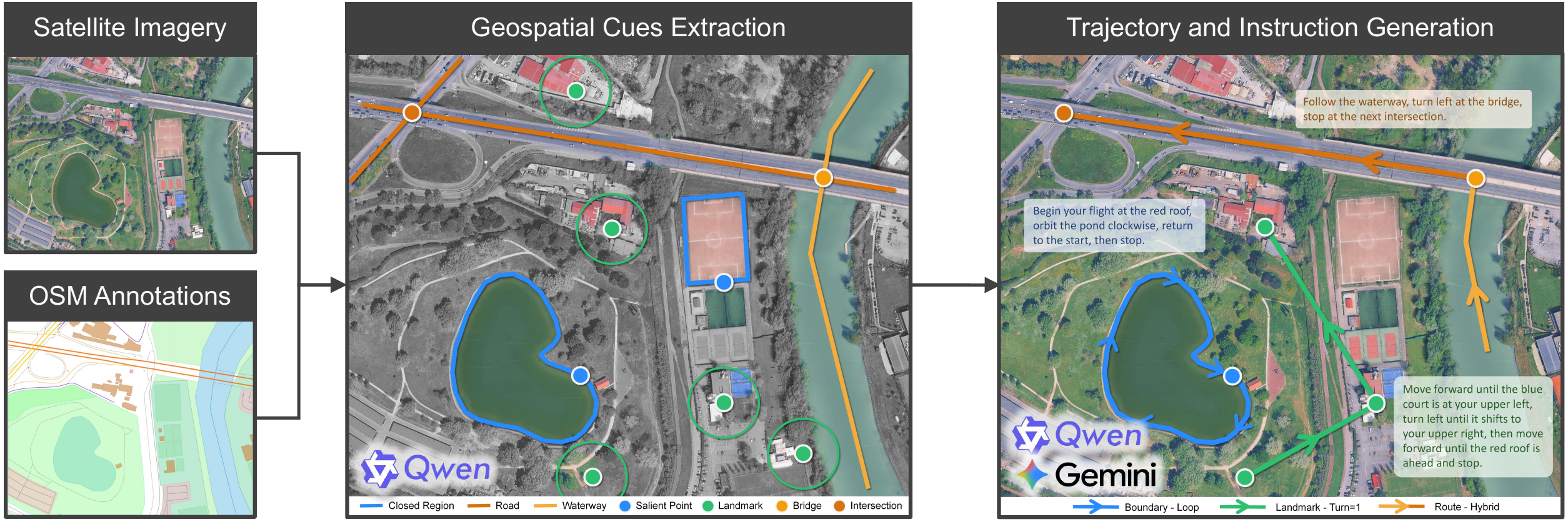}
    \caption{Overview of the SatNav episode-generation pipeline. Structured geospatial cues for the three task families are extracted from aligned satellite imagery and OSM annotations and then used to generate long-horizon trajectories and cue-grounded navigation instructions.}
    \label{fig:route_episode_generation_pipeline}
    \vspace{-10pt}
\end{figure*}

SatNav generates episodes through a unified cue-to-episode pipeline, as illustrated in Figure~\ref{fig:route_episode_generation_pipeline}.
Starting from aligned satellite imagery and OSM annotations, the pipeline extracts validated geospatial cues with basic descriptions, constructs task-specific long-horizon trajectories over these cues, and converts the associated cue sequences into navigation instructions.
Each retained trajectory-instruction pair is then packaged as a standard VLN episode.
This design, detailed in Appendix~\ref{app:Detailed_Pipeline}, allows \textbf{Boundary}, \textbf{Landmark}, and \textbf{Route} episodes to share the same construction pipeline while preserving their distinct evaluative focus.

\vspace{-5pt}
\paragraph{Cue extraction and validation.}
For each selected scene, SatNav aligns the satellite image and OSM annotations in a shared metric coordinate frame.
From this aligned representation, SatNav extracts structured geospatial cues for episode construction.
\textbf{Boundary} episodes use closed-region cues extracted from OSM polygons, with visually salient boundary points selected as candidate start and goal locations. \textbf{Landmark} episodes use visually distinctive local cues detected and localized from satellite crops to ground trajectory waypoints. \textbf{Route} episodes use linear-structure cues by converting roads and waterways into graph nodes and edges with intersection and way-type attributes. Across all three tasks, cues are retained only if they are geometrically stable and visually clear from the relevant satellite observations. For each retained cue, Qwen3.5~\citep{qwen35} generates a basic visual or semantic description, which is later used for instruction construction.

\vspace{-5pt}
\paragraph{Trajectory generation.}
Given the validated cue set, SatNav generates candidate trajectories according to the structure of each task family. For \textbf{Boundary} tasks, trajectories are instantiated by pairing salient boundary points as start and goal locations and tracing the boundary between them. For \textbf{Landmark} tasks, trajectories connect waypoints whose key decisions can be grounded by nearby visual landmarks. For \textbf{Route} tasks, trajectories are sampled as paths over the road-and-waterway graph. The resulting candidates are further filtered and refined using a set of geometric and task-specific rules. For example, we remove geometrically implausible trajectories, such as routes with abrupt large-angle detours, and prioritize long, coherent trajectories so that successful navigation requires agents to use long-term navigation memory.
The retained trajectories are stored as pose sequences.

\vspace{-5pt}
\paragraph{Instruction construction.}
Navigation instructions are grounded in the physical and geometric cues associated with each retained trajectory.
To capture the distinct characteristics of different navigation behaviors, SatNav organizes these cues according to the three task families.
For \textbf{Boundary} episodes, we combine the start and goal boundary points, the closed-region description, the traversal direction, and the loop type.
For \textbf{Landmark} episodes, we order waypoint descriptions along the path and add relative rotation cues at turning waypoints based on the angle between adjacent trajectory segments, e.g., ``turn right until the roofs appear in the lower-right of the view.''
For \textbf{Route} episodes, we serialize the nodes and edges along the trajectory, augmenting each node with its turn direction and ordinal position, e.g., ``at the third bridge, turn left onto the road.''
We then use a two-stage construction pipeline to convert these task-specific cue sequences into diverse navigation instructions.
Qwen~3.5 first composes cue-level descriptions into instruction prototypes in three styles: detailed, natural, and concise.
To reduce within-style homogeneity, Gemini~3 Flash~\citep{gemini2024family} further rewrites these prototypes for linguistic diversity and cross-checks their semantic alignment with the underlying trajectory cues. Further details on instruction prompt templates are provided in Appendix~\ref{app:instruction_generation_prompts}.

\vspace{-5pt}
\paragraph{Episode packaging.} The final stage standardizes the generated data into the unified SatNav episode format. Each episode encapsulates essential navigation components, including the scene identifier, task metadata, navigation instruction, and reference trajectory. These standardized episodes are directly compatible with the SatNav platform, which offers local satellite observations for navigation.

\begin{figure*}[t]
    \centering
    \captionsetup[subfigure]{font=footnotesize, justification=centering}
    \begin{subfigure}[t]{0.27\textwidth}
        \centering
        \includegraphics[width=\linewidth]{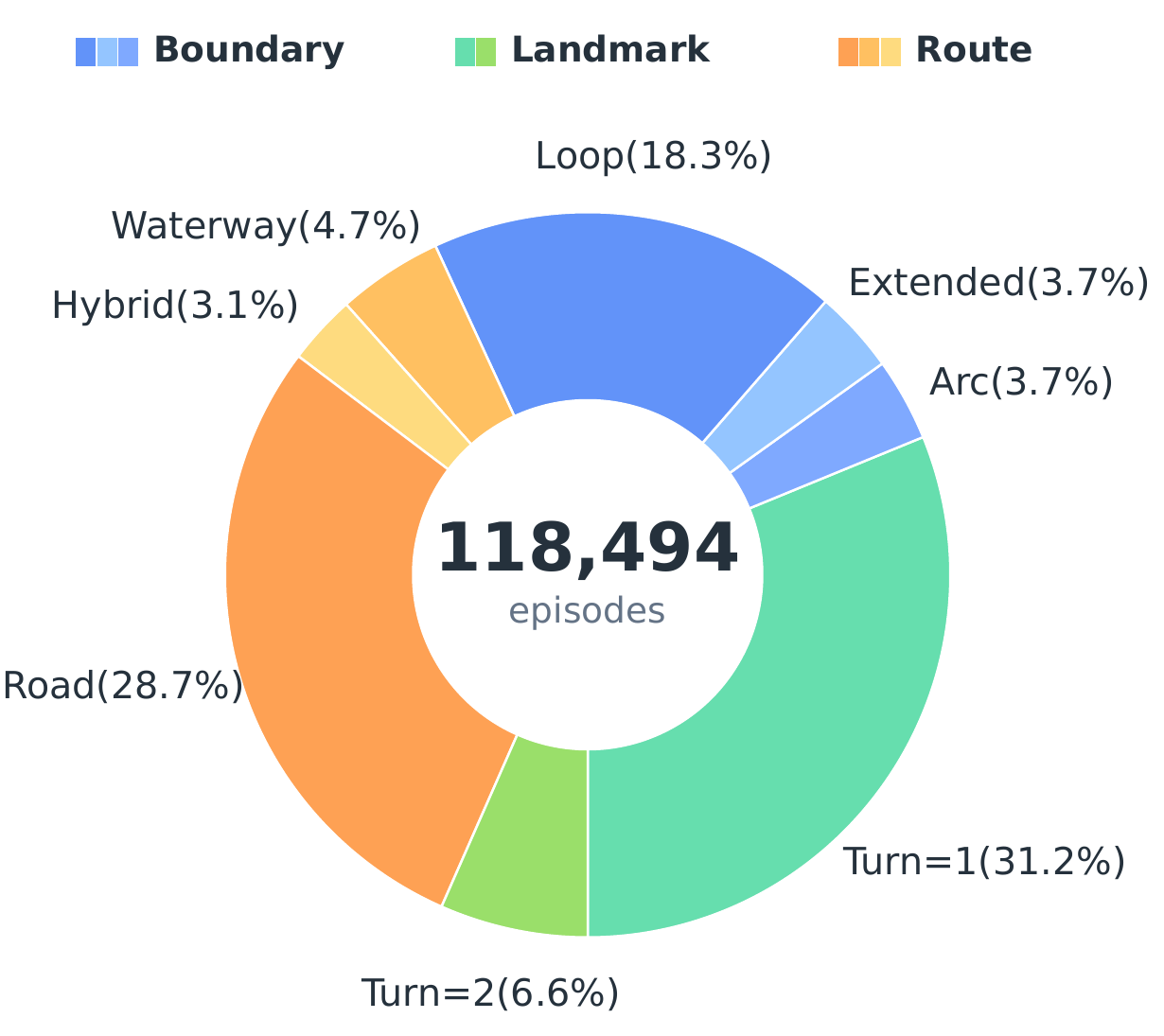}
        \caption{Task subtype distribution}
    \end{subfigure}\hfill
    \begin{subfigure}[t]{0.25\textwidth}
        \centering
        \includegraphics[width=\linewidth]{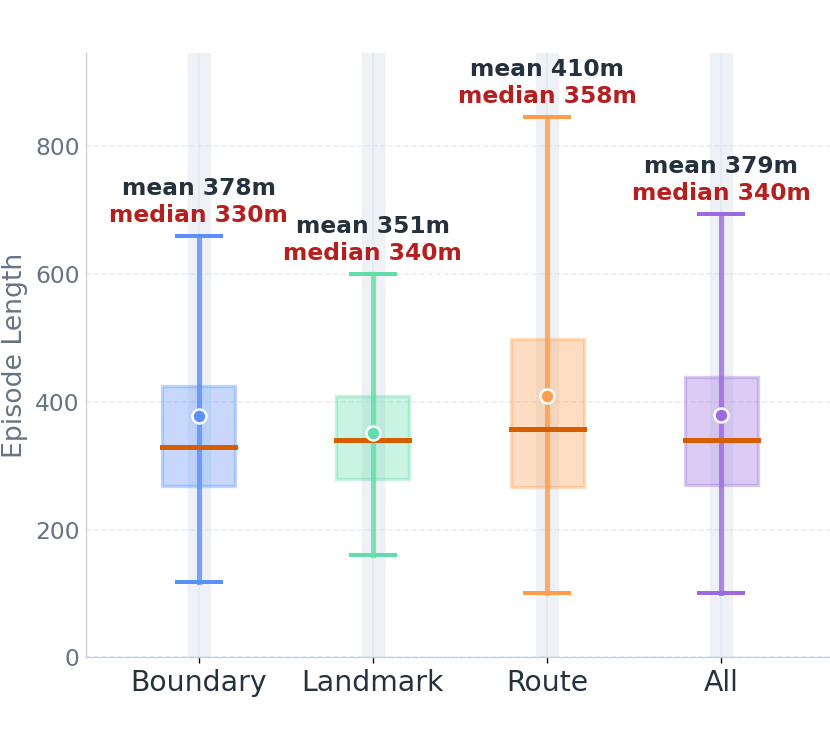}
        \caption{Episode length distribution}
    \end{subfigure}\hfill
    \begin{subfigure}[t]{0.22\textwidth}
        \centering
        \includegraphics[width=\linewidth]{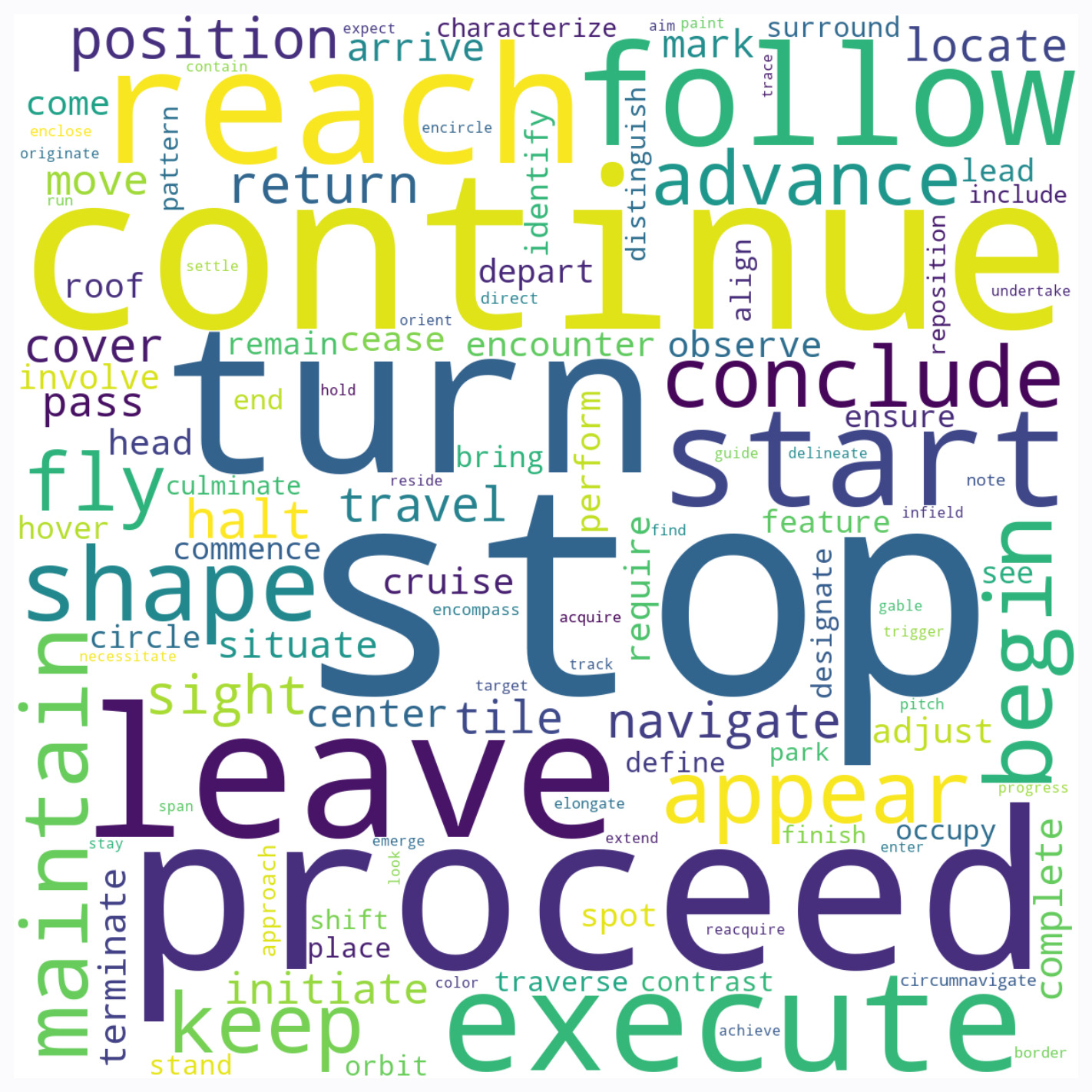}
        \caption{Verb word cloud}
    \end{subfigure}\hfill
    \begin{subfigure}[t]{0.22\textwidth}
        \centering
        \includegraphics[width=\linewidth]{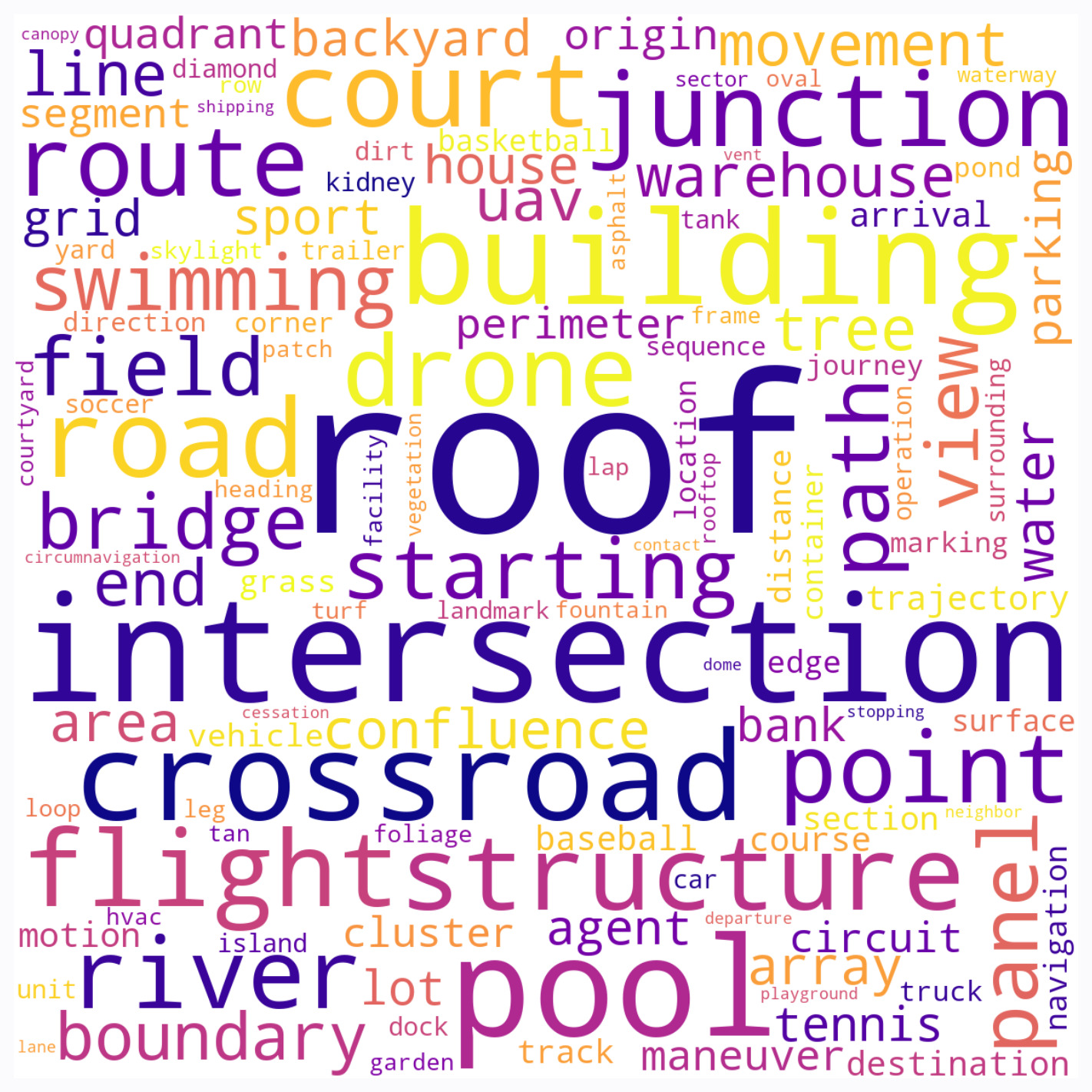}
        \caption{Noun word cloud}
    \end{subfigure}
    \caption{SatNav episode-level statistics}
    \label{fig:episode_statistics}
    \vspace{-10pt}
\end{figure*}

\vspace{-2pt}
\subsection{Dataset Statistics and Splits}
\label{subsec:data_statistics_and_splits}
\vspace{-2pt}
\paragraph{Dataset statistics.}
Figure~\ref{fig:episode_statistics} summarizes the main statistics of SatNav.
The benchmark contains 118,494 episodes constructed from 59 scenes across 18 cities on five continents.
At the episode level, \textbf{Boundary}, \textbf{Landmark}, and \textbf{Route} tasks account for 25.5\%, 38.1\%, and 36.3\% of the dataset, respectively.
The subtype distribution is intentionally non-uniform: common subtypes, such as \textit{full-loop}, \textit{one-turn}, and \textit{road-only} trajectories, provide broad coverage, while less frequent subtypes preserve more challenging cases.
SatNav emphasizes long-horizon navigation, with an average trajectory length of 379\,m and a median length of 340\,m.
As shown in Figure~\ref{fig:episode_statistics}(b), {Route} episodes exhibit the largest variance because distances between key topological features, such as bridges, vary widely across urban layouts.
Linguistically, SatNav contains a 4,740-word vocabulary, with an average instruction length of 41 words.
The word clouds highlight frequent top-down cues, such as roads, intersections, and roofs, reflecting SatNav's distinctive nadir-view aerial perspective.
A comparison of dataset statistics with existing UAV VLN benchmarks is provided in Appendix~\ref{app:benchmark_comparison}.

\vspace{-5pt}
\paragraph{Dataset splits.}
SatNav is organized into Train, Test Seen, and Test Unseen splits. The Train split contains 56 scenes across 15 cities and accounts for 88.7\% of the total data. Test Seen contains 4,574 episodes generated from three training scenes, but uses newly sampled trajectories and newly constructed instructions to evaluate trajectory- and instruction-level generalization in familiar environments. Test Unseen contains 8,756 episodes from three previously unseen scenes in three new cities, and is designed to evaluate generalization to unseen geographic environments. Both test splits maintain the same task distribution, with 33\% Boundary, 33\% Landmark, and 34\% Route episodes, reducing the effect of task imbalance when comparing seen and unseen performance.

\section{SwiftVLN}
\label{sec:swiftvln}

\subsection{Framework Overview}
\label{sec:swiftvln_framework_overview}

\begin{figure}[t]
    \centering
    \includegraphics[width=0.85\textwidth]{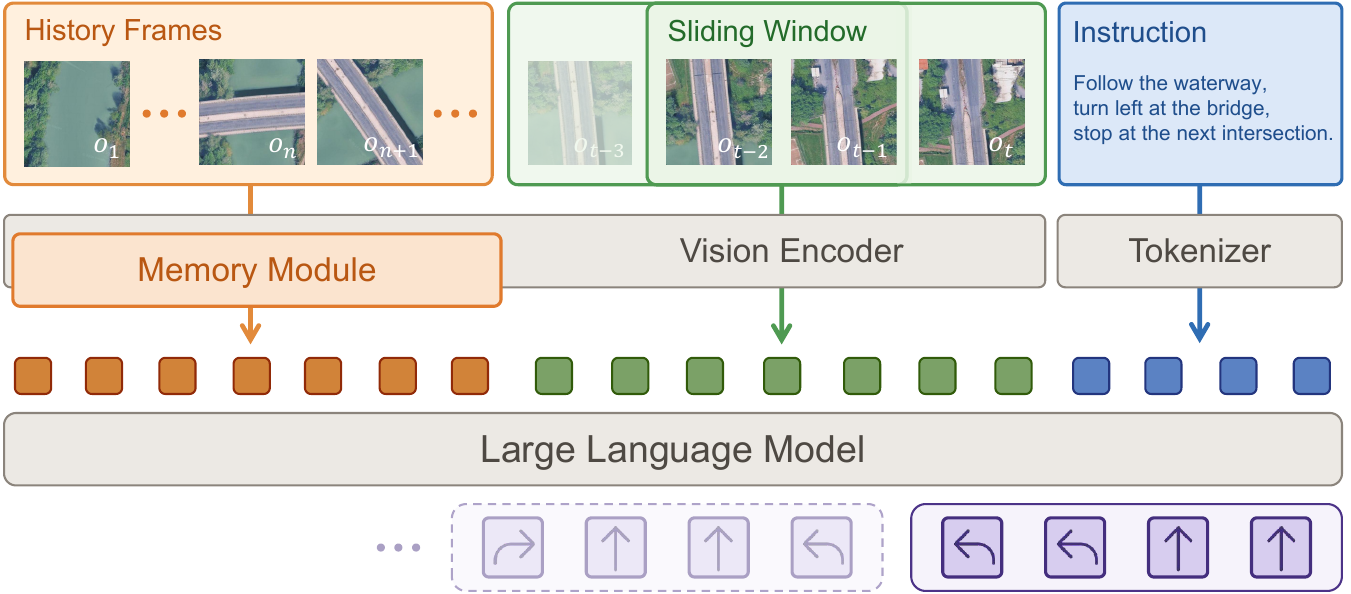}
    \caption{Overview of the SwiftVLN framework.}
    \label{fig:swiftvln_model_arch}
    \vspace{-10pt}
\end{figure}

The SwiftVLN framework refactors the training and evaluation pipeline of StreamVLN~\citep{wei2025streamvln} within ms-swift~\citep{zhao2024swift} into a flexible and extensible VLN structure. This framework enables controlled substitution and evaluation of diverse memory designs, while facilitating the adaptation of different LVLMs to VLN. Figure~\ref{fig:swiftvln_model_arch} provides an overview of this architecture.

Following the dual-memory design of StreamVLN, SwiftVLN maintains both short-term dialogue memory $S_t$ and long-term memory $L_t$ at each model-query round $t$. The short-term dialogue memory $S_t$ is implemented as a multi-turn dialogue window that stores recent image-action turns. To improve its flexibility, SwiftVLN introduces a sliding window mechanism: once the window reaches its maximum capacity $N_w$, it slides forward by removing the oldest turns while retaining $N_o$ overlapping turns in the newly constructed dialogue window. This window sliding mechanism preserves recent context and maintains local continuity across successive windows.

Historical observations that fall outside the short-term context window are processed by a long-term memory module $\mathcal{M}$ to form $L_t$. To facilitate controlled comparisons, SwiftVLN implements $\mathcal{M}$ as a suite of interchangeable mechanisms:
\begin{itemize}[leftmargin=1.5em, labelsep=0.5em, topsep=2pt, itemsep=3pt, parsep=0pt]
    \item \textbf{History Frame Sampling:} Extracts past observations using \textit{Uniform}, \textit{Random}, or \textit{Temporal-Biased} strategies, where the latter explicitly prioritizes recent frames over distant ones.
    
    \item \textbf{Input Augmentation:} \textit{Initial Frame Prompting} adds the first-frame tokens to the system context as a stable visual anchor for the starting scene. \textit{Pose Encoding} augments each visual frame with relative pose cues to provide spatial displacement and heading information.
    
    \item \textbf{Memory Compression:} \textit{Map Memory} replaces raw history frames with constructed global and local maps encoded as a compact memory block. \textit{Global Token Clustering (GTC)} aggregates tokens from all retained frames into a fixed-capacity global memory. To retain chronological order, \textit{Segment Token Clustering (STC)} partitions history into $N$ temporal segments and clusters each segment independently to preserve coarse temporal structure.
    
\end{itemize}

Together, these plug-and-play variants establish a comprehensive testbed for systematically ablating memory architectures. Additional details are provided in Appendices~\ref{app:swiftvln_pipeline} and~\ref{app:swiftvln_memory_details}.

\subsection{Satellite-to-UAV Generalization}
\label{sec:swiftvln_sim2real}
A key concern for SatNav is whether policies trained on satellite imagery can be used with real UAV nadir observations. Although both views are top-down, satellite and UAV images differ in resolution, illumination, altitude, and imaging characteristics, creating a visual domain gap that makes direct deployment unreliable. 
We show that this gap can be bridged with a lightweight adapter and a modest amount of paired data. Using 24K paired UAV-satellite images from cross-view geo-localization datasets~\citep{dai2023vision, ji2025game4loc, zhu2023sues, xu2024uav}, we train a Transformer adapter to map UAV visual tokens into the satellite feature space defined by the frozen vision encoder. The adapter is optimized with a bidirectional contrastive loss and a cosine similarity loss. During inference, the adapter is placed between the vision encoder and the SatNav-trained backbone, enabling the same VLN policy to operate on real UAV nadir images. These results suggest that SatNav-trained policies can transfer to UAV execution with limited adaptation data, supporting the practical relevance of satellite-based scalable VLN training. Additional details are provided in Appendix~\ref{app:swiftvln_sim2real_details}.

\section{Experiments}

\subsection{Experimental Setup}

\subsubsection{Baselines and Ablation Settings}

We evaluate both classical VLN methods (Seq2Seq~\citep{anderson2018vision}, CMA~\citep{wang2019reinforced}) and recent LVLM-based approaches (NaVILA~\citep{cheng2024navila}, UniNaVid~\citep{zhang2024uni}, StreamVLN~\citep{wei2025streamvln}, OpenFly-Agent~\citep{gao2025openfly}) on SatNav. All baselines are adapted to the same setting, using the episode instruction and cropped satellite observations as input at each decision step. While classical methods are trained from scratch, LVLM-based models are initialized from either general-purpose base LVLMs or released navigation-specific checkpoints. All baselines are trained or fine-tuned on the SatNav training split and evaluated on the test splits.

To isolate the effects of different memory designs, we establish a \textbf{SwiftVLN reference model} following the StreamVLN memory configuration: a short-term sliding window of $N_w=8$ with no overlap ($N_o=0$), and a long-term memory constructed by uniformly sampling 8 frames from the trajectory history. Building on this reference model, we conduct memory-module ablations that evaluate both short-term window overlap and the long-term memory variants introduced in Sec.~\ref{sec:swiftvln_framework_overview}, while keeping all non-memory components and training protocols fixed. Detailed training settings and model specifications are provided in Appendix~\ref{app:baseline}.

\subsubsection{Evaluation Metrics}
Following prior VLN works, we report five standard metrics: Success Rate (SR), Oracle Success (OS), Success weighted by Path Length (SPL), Navigation Error (NE), and average Steps. Success is defined as predicting \texttt{stop} within 10\,m of the target, or within 30\,m for the more challenging Landmark task. For Boundary episodes, where the start may coincide with the goal, SR and OS are counted only if the agent first navigates more than 20\,m away from the start before later returning within the success threshold. For episode $i$, we compute $\mathrm{SPL}_i = S_i L_i / \max(L_i, P_i)$, where $S_i$ indicates success, $L_i$ is the reference trajectory length, and $P_i$ is the distance traveled by the agent. NE is the final distance to the goal in meters. Steps counts executed environment actions, including turns and \texttt{stop}. We compute each metric over all episodes in each evaluation split. We report SR, SPL, OS, and percentage-point changes to one decimal place, and NE and average Steps to two decimal places.

\subsection{Baseline Comparison on SatNav}

\begin{table*}[t]
\centering
\caption{Performance comparison on SatNav. SR, SPL, and OS are reported as percentages. Best results are in \textbf{bold}, and second-best results are \underline{underlined}.}
\label{tab:baseline_seen_unseen}
\setlength{\tabcolsep}{4.2pt}
\footnotesize
\resizebox{0.85\textwidth}{!}{%
\begin{tabular}{@{}lcccccccccc@{}}
\toprule
\multirow{2}{*}{Model}
& \multicolumn{5}{c}{Test Seen}
& \multicolumn{5}{c}{Test Unseen} \\
\cmidrule(lr){2-6} \cmidrule(lr){7-11}
& SR$\uparrow$ & SPL$\uparrow$ & OS$\uparrow$ & NE$\downarrow$ & Steps
& SR$\uparrow$ & SPL$\uparrow$ & OS$\uparrow$ & NE$\downarrow$ & Steps \\
\midrule
Seq2Seq    & 2.1  & 2.1  & 31.1 & 180.36 & 53.29 & 1.6  & 1.5  & 29.6 & 219.17 & 58.50 \\
CMA        & 9.5  & 9.3  & 44.9 & 323.25 & 94.85 & 7.3  & 7.2  & 43.9 & 363.59 & 99.31 \\
OpenFly    & 13.2 & 13.0 & 33.5 & 166.88 & 58.93 & 11.7 & 11.6 & 32.2 & 195.91 & 66.35 \\
OpenFly$^\ast$   & 21.1 & 21.0 & 38.1 & 163.07 & 57.85 & 17.1 & 16.9 & 34.9 & 196.90 & 65.56 \\
NaVILA           & 18.1 & 18.0 & 27.6 & 93.05  & 48.08 & 13.0 & 12.7 & 23.7 & 128.88 & 53.71 \\
NaVILA$^\ast$    & 25.0 & 24.9 & 35.0 & 93.88  & 51.47 & 18.6 & 18.4 & 31.6 & 123.49 & 59.57 \\
UniNaVid   & 25.1 & 24.8 & 60.4 & 174.68 & 73.21 & 20.4 & 20.0 & 49.9 & 228.46 & 80.55 \\
UniNaVid$^\ast$  & 49.7 & 49.1 & 68.2 & 87.11  & 53.99 & 36.7 & 36.3 & 55.9 & 149.85 & 64.75 \\
StreamVLN  & 64.3 & 63.7 & {71.7} & 53.24 & 51.58 & 52.2 & {51.8} & 61.0 & \underline{84.99} & 57.80 \\
StreamVLN$^\ast$ \ \ \ \  & \textbf{67.3} & \textbf{66.8} & \textbf{74.6} & \underline{47.47} & {52.11} & \textbf{58.4} & \textbf{57.8} & \textbf{68.3} & {86.63} & 61.19 \\
\midrule
\textbf{SwiftVLN} & \underline{65.8} & \underline{65.5} & \underline{72.4} & \textbf{34.05} & 49.97 & \underline{53.7} & \underline{53.2} & \underline{64.1} & \textbf{62.29} & 57.38 \\
\bottomrule
\end{tabular}%
}
\vspace{2pt}
\begin{minipage}{\textwidth}
\vspace{2pt}
\footnotesize
$^\ast$ indicates models fine-tuned on SatNav from released checkpoints trained on each model's original navigation task. Unmarked models are initialized from their corresponding base backbones.
\end{minipage}
\vspace{-13pt}
\end{table*}

Table~\ref{tab:baseline_seen_unseen} summarizes baseline performance on SatNav.
Overall, performance drops consistently from Test Seen to Test Unseen, highlighting the generalization challenges in novel geographic regions.
Across most methods, SR and SPL are close, indicating that successful agents typically follow paths with limited detours, while large deviations remain difficult to recover from in open satellite-map environments.
Classical VLN baselines (Seq2Seq and CMA) substantially lag behind LVLM-based methods, likely because they were designed for egocentric RGB-D settings rather than top-down observations.
CMA obtains relatively high OS but much lower SR, suggesting that it can pass near the target but often fails to stop reliably.
Navigation-pretrained initialization consistently improves performance.
This is particularly evident in UniNaVid, where the continued variant improves SR by 24.6 and 16.3 percentage points on Test Seen and Test Unseen, respectively, suggesting that navigation-specific pretraining transfers useful temporal and action priors to SatNav tasks.
However, OpenFly$^\ast$ substantially underperforms other LVLM baselines despite being initialized from an outdoor UAV checkpoint.
This underperformance likely stems from its restricted visual history and a mismatch between its inherited multi-dimensional control priors and SatNav's discrete action space.
StreamVLN$^\ast$ achieves the highest SR, SPL, and OS. It uses a navigation-pretrained 7B model, whereas SwiftVLN starts from the general-purpose Qwen2.5-VL-3B backbone. SwiftVLN achieves the lowest NE on both splits and provides a modular framework for controlled memory ablations.

\vspace{-2pt}
\subsection{Memory Design Analysis}
\vspace{-2pt}

We conduct memory-design ablations on the SwiftVLN reference model, as shown in Table~\ref{tab:ablation_memory}.
Removing long-term memory leads to a clear performance drop, showing that short-term history alone is not sufficient for SatNav tasks. 
Short-term window overlap improves over the non-overlap reference setting, suggesting that the local continuity preserved by sliding windows benefits long-horizon decision making.
For history-frame sampling, random sampling performs poorly, while temporal-biased sampling brings small but consistent gains.
This is likely because recent observations are often more informative for the next action, while a smaller number of earlier frames can still serve as anchors for the past trajectory.

Input augmentation gives mixed results.
Adding the initial observation improves OS but not SR, suggesting that it helps the agent pass near the goal while still failing to stop reliably.
Relative pose encoding brings stable gains, indicating that explicit spatial offsets improve frame interpretation.
For memory compression, map memory and GTC both underperform the reference model.
The degradation in map memory likely stems from our current implementation, which shares a single vision encoder for explored maps and raw observations.
This shared parameter space may force a representation compromise, showing that a dedicated map encoder may be needed.
STC outperforms GTC by compressing history within temporal segments and preserving coarse temporal order.

\begin{table*}[t]
\centering
\caption{SwiftVLN ablation study of memory designs on SatNav. SR, SPL, and OS are reported as percentages, while $\Delta$ columns report absolute percentage-point changes relative to SwiftVLN.}
\label{tab:ablation_memory}
\setlength{\tabcolsep}{5.0pt}
\footnotesize
\resizebox{\textwidth}{!}{%
\begin{tabular}{@{}lcccccccccc@{}}
\toprule
\multirow{2}{*}{Memory Design}
& \multicolumn{5}{c}{Test Seen}
& \multicolumn{5}{c}{Test Unseen} \\
\cmidrule(lr){2-6} \cmidrule(lr){7-11}
& SR$\uparrow$ & SPL$\uparrow$ & OS$\uparrow$ & $\Delta$SR$\uparrow$ & $\Delta$OS$\uparrow$
& SR$\uparrow$ & SPL$\uparrow$ & OS$\uparrow$ & $\Delta$SR$\uparrow$ & $\Delta$OS$\uparrow$ \\
\midrule
\multicolumn{11}{@{}l}{\textbf{\textit{Memory necessity}}} \\
SwiftVLN reference              & 65.8 & 65.5 & 72.4 & 0.0 & 0.0 & 53.7 & 53.2 & 64.1 & 0.0 & 0.0 \\
Short-term only         & 44.5 & 43.8 & 62.8 & -21.3 & -9.6 & 32.4 & 31.5 & 51.4 & -21.3 & -12.7 \\
\midrule
\multicolumn{11}{@{}l}{\textbf{\textit{Short-term sliding window}}} \\
Overlap turns($N_o=2$) & 68.7 & 64.0 & 81.3 & +2.9 & \textbf{+8.9} & 56.8 & 53.7 & 70.1 & +3.1 & +6.0 \\
Overlap turns($N_o=4$) & 71.3 & 70.9 & 80.6 & \textbf{+5.5} & +8.2 & 60.3 & 59.9 & 71.5 & \textbf{+6.6} & \textbf{+7.4} \\
\midrule
\multicolumn{11}{@{}l}{\textbf{\textit{History-frame sampling}}} \\
Random sampling         & 52.0 & 51.6 & 67.9 & -13.8 & -4.5 & 41.1 & 40.9 & 57.6 & -12.6 & -6.5 \\
Temporal-biased sampling & 66.9 & 66.4 & 75.8 & \textbf{+1.1} & \textbf{+3.4} & 55.8 & 55.3 & 66.3 & \textbf{+2.1} & \textbf{+2.2} \\
\midrule
\multicolumn{11}{@{}l}{\textbf{\textit{Input augmentation}}} \\
+ Initial observation  & 62.7 & 61.6 & 77.5 & -3.1 & \textbf{+5.1} & 52.6 & 51.7 & 67.5 & -1.1 & \textbf{+3.4} \\
+ Relative pose         & 67.8 & 67.5 & 76.3 & \textbf{+2.0} & +3.9 & 55.4 & 55.2 & 65.0 & \textbf{+1.7} & +0.9 \\
\midrule
\multicolumn{11}{@{}l}{\textbf{\textit{Long-term Memory compression}}} \\
Map memory              & 59.9 & 59.6 & 73.4 & -5.9 & +1.0 & 48.7 & 48.4 & 63.1 & -5.0 & -1.0 \\
Global token clustering (GTC) & 63.5 & 63.2 & 72.7 & -2.3 & +0.3 & 51.4 & 51.0 & 62.4 & -2.3 & -1.7 \\
Segment token clustering (STC)\ \ \ \ \ \ \ \  \ \ \ \  & 68.3 & 67.9 & 77.3 & \textbf{+2.5} & \textbf{+4.9} & 54.5 & 54.1 & 66.2 & \textbf{+0.8} & \textbf{+2.1} \\
\bottomrule
\end{tabular}%
}
\end{table*}

\vspace{-2pt}
\subsection{Real-World Experiments}
\begin{figure}
  \begin{subfigure}{\textwidth}
    \centering
    \includegraphics[width=\textwidth,page=1]{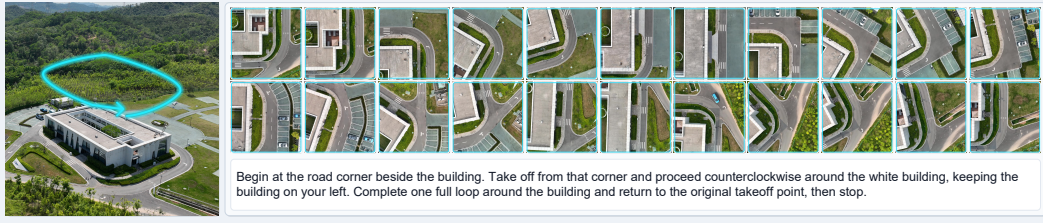}
  \end{subfigure}
  \caption{A real-world \textbf{Boundary} task around a building with a grey roof.}
  \label{fig:traj_true_building}
  \vspace{-10pt}
\end{figure}

\vspace{-2pt}
\label{sec:real_world_demo}
We use a DJI Matrice 4D UAV for real-world evaluation. Nadir-view images are transmitted via a video link to a local workstation, where SwiftVLN performs online inference on an NVIDIA RTX 4090 GPU.
We construct 18 real-world episodes spanning all three SatNav task families, with six episodes per family. Instructions are initially written by humans and subsequently refined using the same LLM-based rewriting pipeline employed in SatNav to maintain consistency with the dataset's instruction style. Figure~\ref{fig:traj_true_building} illustrates a representative \textit{Boundary} episode in which the UAV follows the perimeter of a white building. Table~\ref{tab:real_world_results} reports the real-world navigation performance of the SwiftVLN reference model with and without the visual adapter.

These flights demonstrate the feasibility of deploying a satellite-trained navigation model on a real UAV with lightweight visual adaptation. However, real-world conditions remain challenging, particularly in cases involving local scene ambiguity and execution uncertainty, highlighting room for further improvement in satellite-to-UAV transfer. Details of the experimental setup, additional qualitative results, and analyses of representative cases are provided in Appendix~\ref{app:real_world_demo}.

\begin{table*}[t]
\centering
\caption{Real-world navigation performance of SwiftVLN with and without the visual adapter. Each setting is evaluated on 18 episodes, with six per task family. SR is reported as a percentage.}
\label{tab:real_world_results}
\setlength{\tabcolsep}{5pt}
\renewcommand{\arraystretch}{1.15}
\begin{tabular}{lcccc}
\toprule
Method & Boundary SR & Landmark SR & Route SR & Overall SR \\
\midrule
SwiftVLN & 33.3 & 33.3 & 50.0 & 38.9 \\
SwiftVLN w/ adapter & 33.3 & 50.0 & 83.3 & 55.6 \\
\bottomrule
\end{tabular}
\end{table*}

\section{Limitations}
SatNav has three main limitations.
First, navigation uses four discrete actions at a fixed altitude. We generate trajectories using these actions and remove paths that require other movements. This favors smooth routes and limits coverage of winding paths. Extending the action space to include altitude and speed changes, takeoff, and landing would support a wider range of UAV missions.
Second, the current training pipeline does not support reinforcement fine-tuning. Training on reference trajectories provides limited experience with states reached after an agent makes a mistake. Adding training through interaction could help agents learn to recover from wrong turns and improve stopping decisions over long routes.
Third, system latency limits the number of long-duration real-flight trials. Our evaluation on 18 episodes provides preliminary evidence of satellite-to-UAV transfer. The current transfer method also uses a simple visual adapter. Future work can design and evaluate more effective methods for satellite-to-UAV transfer.

\section{Conclusion}
We presented \textbf{SatNav}, a scalable benchmark for evaluating memory-intensive long-horizon UAV VLN from satellite imagery.
By constructing episodes directly from satellite imagery and OSM annotations, SatNav avoids reliance on reconstructed 3D assets and provides a scalable path to city-scale benchmark construction.
Through three task families, \textbf{Boundary}, \textbf{Landmark}, and \textbf{Route}, SatNav turns long-range progress tracking, geospatial grounding, and route reasoning into explicit evaluation targets.
Our experiments show that existing classical and LVLM-based navigation agents still face clear challenges in generalizing to unseen geographic environments and stopping reliably.
The memory ablation study on \textbf{SwiftVLN} provides a comparative view of long-horizon memory designs and offers guidance for future VLN model design.
Our real-world demonstrations further show the feasibility of deploying satellite-trained navigation models on UAVs, while also highlighting remaining challenges in satellite-to-UAV transfer.
We hope SatNav will serve as a practical foundation for developing and stress-testing long-horizon aerial navigation agents at urban scale.

\begin{ack}
This work was supported in part by the Guangdong Basic and Applied Basic Research Foundation under Grant 2023A1515111151, the Guangzhou Municipal District Science and Technology Bureau under Grant 2025A03J3655, GDST under Grant C\_2025\_017, BYD under Grant CP2025O007, and research funding under Grant 2023QN10X121.

We thank the Low-Altitude Intelligent Integrated Test Base in Longgang, Shenzhen, China, for providing the site for our real-flight experiments. We also thank Ziyi Chen from Galbot for discussions during the early stages of this project, and Shuo Sun and Lei Zhang from the Low Altitude Space Economy Research Center (LASER), International Digital Economy Academy (IDEA), for their help with figure design and real-flight experiments. We thank Chao Wang and Jiayang Sun from LASER, IDEA, for maintaining computing resources and helping resolve issues during their use.
\end{ack}

\clearpage
\bibliographystyle{plainnat}
\bibliography{refs}

\newpage
\appendix

\section{Scene Selection and Observation Generation}
\label{app:scene_details}
\subsection{Observation Cropping}
\label{app:observation_cropping}
SatNav uses ego-aligned satellite crops as visual observations.
At step $t$, the agent pose is $p_t=(\lambda_t,\phi_t,\theta_t)$, where $\lambda_t$ and $\phi_t$ denote longitude and latitude in WGS84 (EPSG:4326), and $\theta_t$ denotes heading.
To abstract UAV navigation as planar motion, we fix the observation footprint and remove altitude from the controllable state space.

To generate the observation $o_t$, we account for the difference between the pose coordinate system and the map coordinate system.
Satellite maps are stored in Web Mercator (EPSG:3857), which introduces latitude-dependent scale distortion.
For each pose, we first project $(\lambda_t,\phi_t)$ from WGS84 to Web Mercator.
We then apply the local scale factor $k=1/\cos(\phi_t)$ to correct this distortion.
Specifically, we define an ego-aligned sampling frame centered at the projected location, with its vertical axis aligned to the agent heading $\theta_t$.
Within this rotated frame, we sample a square region with side length $100k$ in Web Mercator meters, corresponding to a true ground footprint of $100\,\mathrm{m}\times100\,\mathrm{m}$.

This fixed coverage matches the nadir-view setting of a UAV flying at $50\,\mathrm{m}$ altitude with a $90^\circ$ horizontal field of view.
Because the sampling frame is already ego-aligned, the resulting crop has the agent's forward direction pointing upward in the image.
The crop is then resized to $448\times448$ pixels, ensuring a consistent metric scale across different cities and latitudes.

\subsection{Scene Collection}
\label{app:scene_collection}
\begin{table*}[t]
\centering
\caption{Statistics of scenes and episodes in the SatNav dataset.}
\label{tab:scene_list}
\small
\resizebox{0.95\textwidth}{!}{%
\begin{tabular}{lccc@{\hspace{0.8cm}}lccc}
\toprule
\textbf{City/Region} & \textbf{Scenes} & \textbf{Avg. Area} & \textbf{Episodes}
& \textbf{City/Region} & \textbf{Scenes} & \textbf{Avg. Area} & \textbf{Episodes} \\
& \textbf{(\#)} & \textbf{(km$^2$)} & \textbf{(\#)}
& & \textbf{(\#)} & \textbf{(km$^2$)} & \textbf{(\#)} \\
\midrule
Amsterdam (EU)     & 2 & 6.56  & 6,424  & Minneapolis (NA)    & 6 & 11.52 & 10,523 \\
Auckland (OC)      & 1 & 15.17 & 2,391  & New York (NA)        & 5 & 16.42 & 10,519 \\
Berlin (EU)        & 5 & 11.58 & 8,942  & Orlando (NA)         & 1 & 47.73 & 3,277  \\
Boston (NA)        & 5 & 14.14 & 11,954 & Paris (EU)           & 5 & 11.55 & 7,892  \\
Brugge (EU)        & 1 & 8.75  & 2,845  & Rio de Janeiro (SA)  & 1 & 22.24 & 1,975  \\
Dubai (AS)         & 1 & 18.97 & 600    & Rome (EU)            & 5 & 12.84 & 8,244  \\
Geneva (EU)        & 5 & 10.62 & 6,342  & Rotterdam (EU)       & 1 & 12.18 & 3,088  \\
London (EU)        & 4 & 11.30 & 5,759  & Sydney (OC)          & 1 & 17.71 & 1,848  \\
Los Angeles (NA)   & 4 & 15.87 & 9,669  & The Bay Area (NA)    & 6 & 15.55 & 16,202 \\
\bottomrule
\end{tabular}%
}
\end{table*}

SatNav is built from city-scale satellite imagery and aligned OSM annotations. We use satellite imagery at tile zoom level 19 (approximately 0.3\,m per pixel), which is suitable for low-altitude UAV navigation and preserves fine-grained semantic detail. Each scene covers about 14\,km$^2$ on average, providing sufficient spatial extent for long-horizon trajectory generation while keeping crop-based observation rendering efficient during training and evaluation.

Scenes are selected using fixed dataset-level criteria rather than task-specific manual design. We retain regions with visually interpretable imagery, usable OSM annotations, and sufficient geospatial diversity for constructing Boundary, Landmark, and Route episodes. The selected scenes span residential, commercial, industrial, park, coastal, riverine, and mixed urban areas. The final benchmark contains 59 scenes across 18 cities; detailed statistics are reported in Table~\ref{tab:scene_list}.

\section{UAV Benchmark Comparison}
\label{app:benchmark_comparison}
\begin{table*}[t]
    \centering
    \small
    \setlength{\tabcolsep}{4pt}
    \begin{threeparttable}
    \caption{Comparison of aerial navigation benchmarks in terms of scale and diversity. Coverage area is reported in $\mathrm{km}^2$, and total trajectory length is reported in $\mathrm{km}$. Best results are in \textbf{bold}, and second-best results are \underline{underlined}.}
    \label{tab:benchmark_comparison}
    \begin{tabular}{lcccccccc}
    \toprule
    & \multicolumn{5}{c}{Scale} & \multicolumn{3}{c}{Diversity} \\
    \cmidrule(lr){2-6} \cmidrule(lr){7-9}
    Dataset & Cov. ($\mathrm{km}^2$) & $N_{\text{scene}}$ & $N_{\text{ep}}$ & Traj. Sum ($\mathrm{km}$) & Avg. Len. & Traj. Types & Vocab. & Lex. Div. \\
    \midrule
    AVDN       & \underline{642} & -- & 3.1K & 879   & 289 & 2 & 2.2K & 0.267 \\
    OpenFly    & 258 & 14 & \underline{100K} & 9.91K & 99  & 1 & \underline{6.4K} & 0.247 \\
    AerialVLN  & --  & \underline{25} & 25.3K & 16.77K & \textbf{662} & 1 & \textbf{7.0K} & 0.246 \\
    HUGE-Bench & 6.45 & 4 & 6.2K & 2.56K & 415 & \textbf{8} & 346 & 0.087 \\
    OpenUAV    & -- & 22 & 12.1K & 3.10K & 255 & 1 & -- & -- \\
    CityNav    & 4.65 & 2 & 32.6K & \underline{17.80K} & \underline{545} & 1 & 4.1K & \textbf{0.354} \\
    UAV-ON     & 9 & 14 & 11.4K & 648 & 57 & 1 & 2.1K & 0.229 \\
    \midrule
    Ours       & \textbf{813} & \textbf{59} & \textbf{118.5K} & \textbf{44.95K} & 379 & \textbf{8} & 4.7K & \underline{0.316} \\
    \bottomrule
    \end{tabular}
    \begin{tablenotes}
    \footnotesize
    \item `--` indicates that the corresponding statistic is not explicitly reported or not reliably available from the released data.
    \end{tablenotes}
    \end{threeparttable}
\end{table*}

Table~\ref{tab:benchmark_comparison} compares SatNav with representative aerial-navigation benchmarks in terms of scale and diversity. The first six statistics are taken from the corresponding original papers whenever available, whereas vocabulary size and lexical diversity are recomputed from the publicly released instruction texts under a unified protocol. Specifically, vocabulary size counts the number of distinct lowercased word forms in the released instructions, without stemming or lemmatization. Lexical diversity is measured by masked Distinct-2~\citep{li2016diversity}: we first normalize instructions and mask route-dependent slots, such as numbers, directional expressions, action words, route markers, and landmark or structure names, and then compute the ratio of unique bigrams to all bigram occurrences. This gives a route-controlled measure of phrase-level lexical variety.

SatNav achieves the strongest overall scale among the compared benchmarks. The results show that SatNav expands aerial VLN across multiple dimensions, and this scalability mainly comes from its use of satellite imagery and map annotations, rather than manually reconstructed 3D assets.
SatNav also provides strong trajectory diversity. SatNav includes eight trajectory types, matching the highest trajectory-type diversity among the compared benchmarks and exceeding most prior aerial VLN and ObjNav datasets, which typically focus on one or two trajectory patterns. In terms of language statistics, SatNav has a vocabulary size of 4.7K and a lexical diversity score of 0.316, indicating broad lexical coverage and relatively diverse instructions under the unified route-controlled protocol.

\section{Automatic Episode Generation Details}
\label{app:Detailed_Pipeline}

This section provides the implementation details of SatNav episode generation. The pipeline is designed to produce high-quality VLN episodes from satellite imagery and OSM-derived geospatial cues through a fixed automatic procedure. Across all scenes and cities, the same processing stages are applied: geospatial cue extraction, quality control, trajectory instantiation, instruction construction, and episode packaging.

\begin{figure*}[t]
    \centering
    \captionsetup[subfigure]{font=footnotesize, justification=centering}

    \begin{subfigure}[t]{0.35\textwidth}
        \centering
        \includegraphics[width=\linewidth]{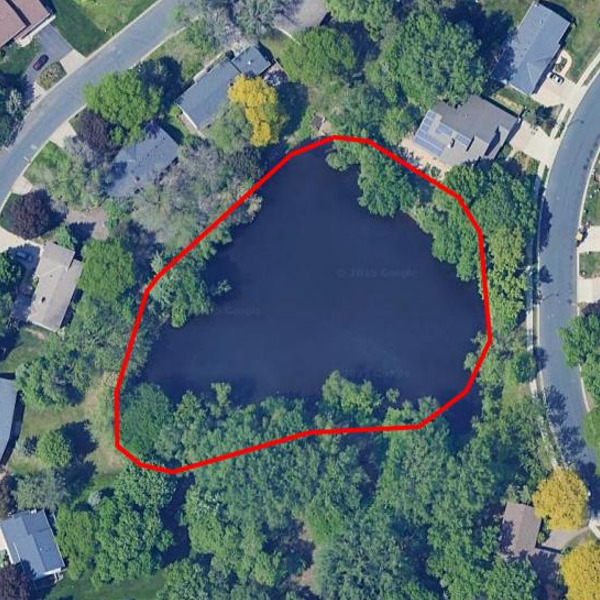}
        \caption{Original OSM boundary}
    \end{subfigure}
    \hspace{2cm}
    \begin{subfigure}[t]{0.35\textwidth}
        \centering
        \includegraphics[width=\linewidth]{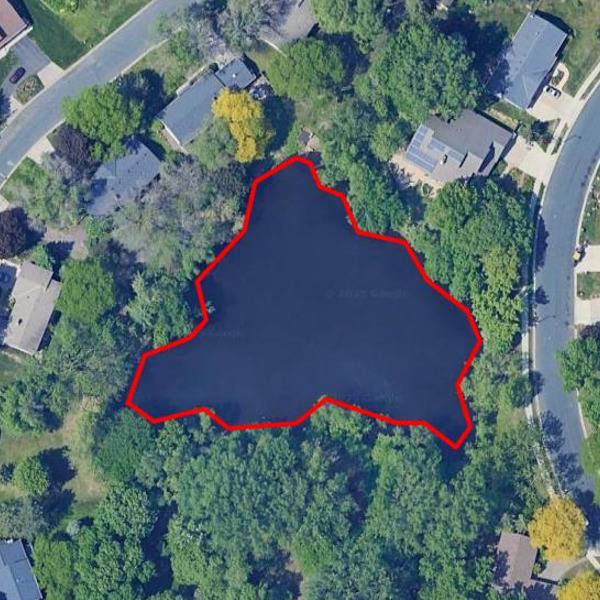}
        \caption{Refined image-aligned boundary (SAM)}
    \end{subfigure}

    \caption{Boundary annotation refinement.}
    \label{fig:boundary_sam}

    \vspace{-10pt}
\end{figure*}

\subsection{Detailed Boundary Pipeline}
\label{app:Detailed_Boundary_Pipeline}

Boundary episodes are constructed from closed geographic structures, such as lakes, stadiums, building complexes, industrial regions, and small islands. We first retrieve OSM polygons from categories that commonly define enclosed regions, including \texttt{building}, \texttt{landuse}, \texttt{leisure}, \texttt{natural}, \texttt{amenity}, and \texttt{place=islet}. Since OSM polygons and satellite imagery may have residual georegistration differences, the polygon is used as an automatic prompt for image-based mask refinement rather than as the final boundary. Specifically, the polygon bounding box and an interior point are provided to SAM~\citep{ravi2024sam}, and the highest-confidence mask is retained when it satisfies the quality-control criteria in Table~\ref{tab:episode_generation_quality_control}. Figure~\ref{fig:boundary_sam} illustrates this refinement process.

\begin{table*}[t]
\centering
\small
\renewcommand{\arraystretch}{1.5} %
\caption{Automatic quality-control criteria used during SatNav episode generation.}
\label{tab:episode_generation_quality_control}

\begin{tabularx}{\textwidth}{
    >{\RaggedRight\arraybackslash}p{2.0cm} %
    >{\hsize=0.7\hsize\RaggedRight}X    %
    >{\hsize=1.3\hsize\RaggedRight}X    %
}
\toprule
\textbf{Stage} & \textbf{Criterion} & \textbf{Purpose} \\
\midrule
Boundary cue \newline extraction & Boundary scale, regularity, interior consistency, and cross-boundary contrast & Retain closed structures that form coherent visible regions and provide clear perimeter-following cues. \\

Landmark cue \newline extraction & Coarse-to-fine visual localization and local-view verification & Retain landmarks that can be localized from satellite crops and associated with nearby decision poses. \\

Route graph \newline construction & Topological validity and visual observability of linear structures & Retain road and waterway segments that are sufficiently visible and can support unambiguous route-following instructions. \\
Trajectory \newline instantiation & Visibility, smoothness, and action-space executability & Remove trajectories whose key cues are outside the local observation, whose turns are ambiguous or visually weak, or whose geometry cannot be followed by the discrete actions. \\

Instruction \newline construction & Cue-trajectory consistency and language validation & Ensure that generated instructions refer only to cues that appear along the executable trajectory and remain semantically consistent after paraphrasing. \\
\bottomrule
\end{tabularx}
\end{table*}

After boundary refinement, candidate anchor points are uniformly sampled along the perimeter and verified by Qwen~3.5. Anchor points are retained when they are associated with nearby visually distinctive cues, such as building corners, roofs, entrances, roads, banks, or water-land transitions. These localized cues are used to construct start and target descriptions, while the refined boundary provides the executable reference path. Depending on the start-target relation along the closed curve, the resulting episodes are categorized as partial-arc, full-loop, or extended-loop boundary tasks.

\subsection{Detailed Landmark Pipeline}
\label{app:Detailed_Landmark_Pipeline}

Landmark episodes are designed to evaluate landmark-grounded turning and relative spatial reasoning. We detect candidate landmarks using a coarse-to-fine localization process over satellite imagery. A coarse scene crop first provides global context for identifying visually distinctive objects or regions, and a finer grid is then used to verify the selected landmark and estimate its location more accurately. This hierarchy keeps the VLM input interpretable while allowing landmarks to be localized over long-range trajectories.

Trajectory waypoints are generated around the retained landmarks under two executability constraints. First, decision waypoints are placed within 30\,m of their associated landmarks so that the cue remains visible in the local observation. Second, segment lengths and turning angles are quantized to the SatNav action space, where the agent moves forward by a fixed metric step and rotates by a fixed angular increment. These constraints ensure that the generated demonstration can be exactly executed by the same action interface used during model evaluation. Instructions are then generated by ordering the landmarks along the trajectory and expressing each key decision through relative-view cues, such as turning until a landmark appears in a specified region of the observation.

\subsection{Detailed Route Pipeline}
\label{app:Detailed_Road_Pipeline}

\begin{figure*}[t]
    \centering
    \captionsetup[subfigure]{font=footnotesize, justification=centering}

    \begin{subfigure}[t]{0.23\textwidth}
        \centering
        \includegraphics[width=\linewidth]{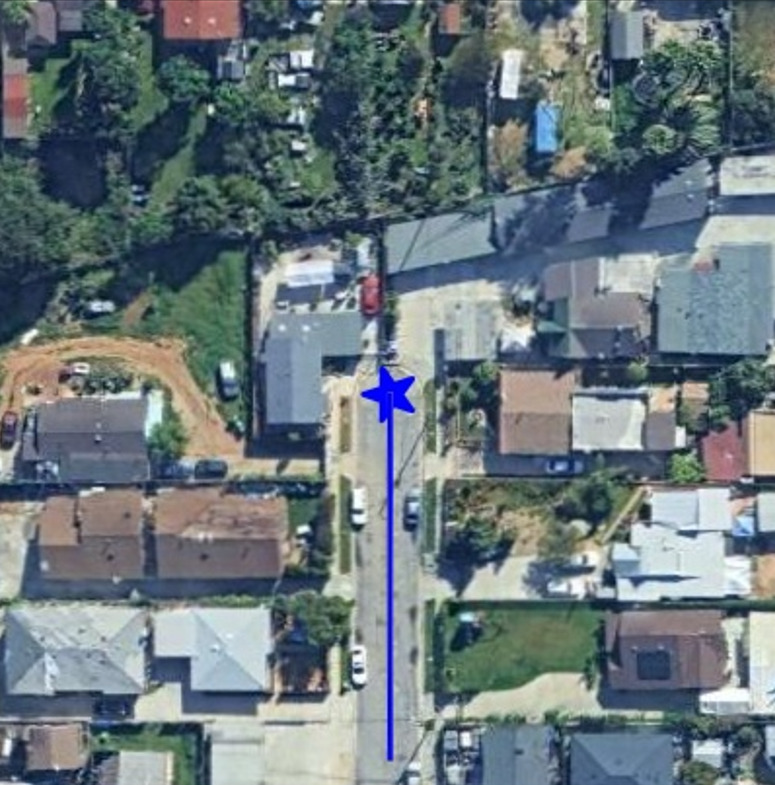}
        \caption{Road dead end}
    \end{subfigure}\hfill
    \begin{subfigure}[t]{0.23\textwidth}
        \centering
        \includegraphics[width=\linewidth]{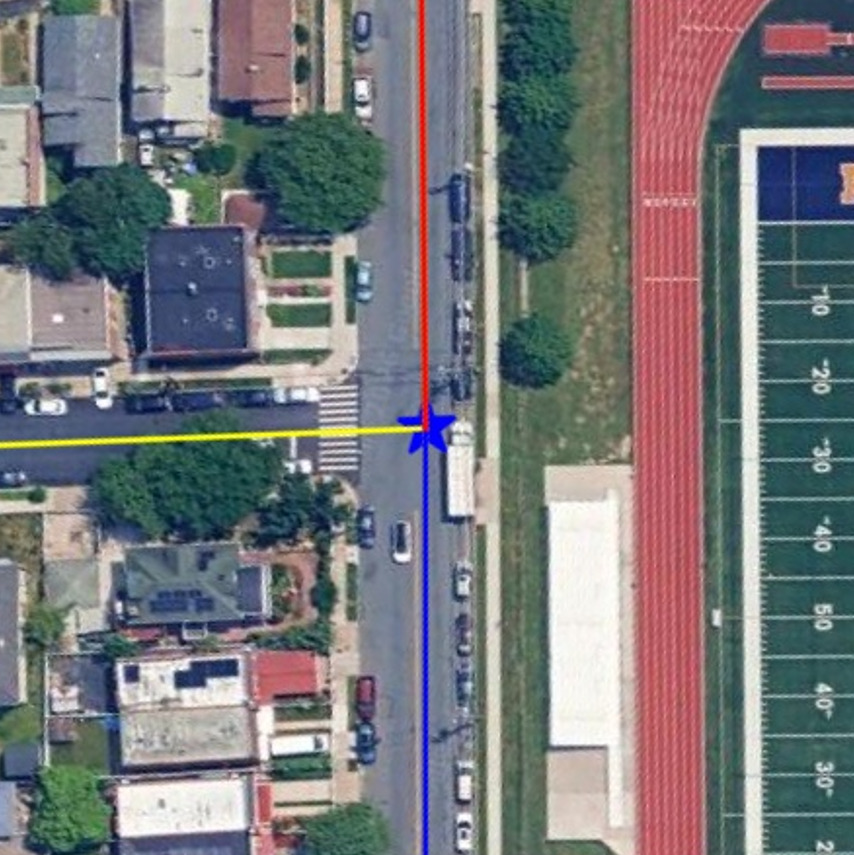}
        \caption{T-shaped intersection}
    \end{subfigure}\hfill
    \begin{subfigure}[t]{0.23\textwidth}
        \centering
        \includegraphics[width=\linewidth]{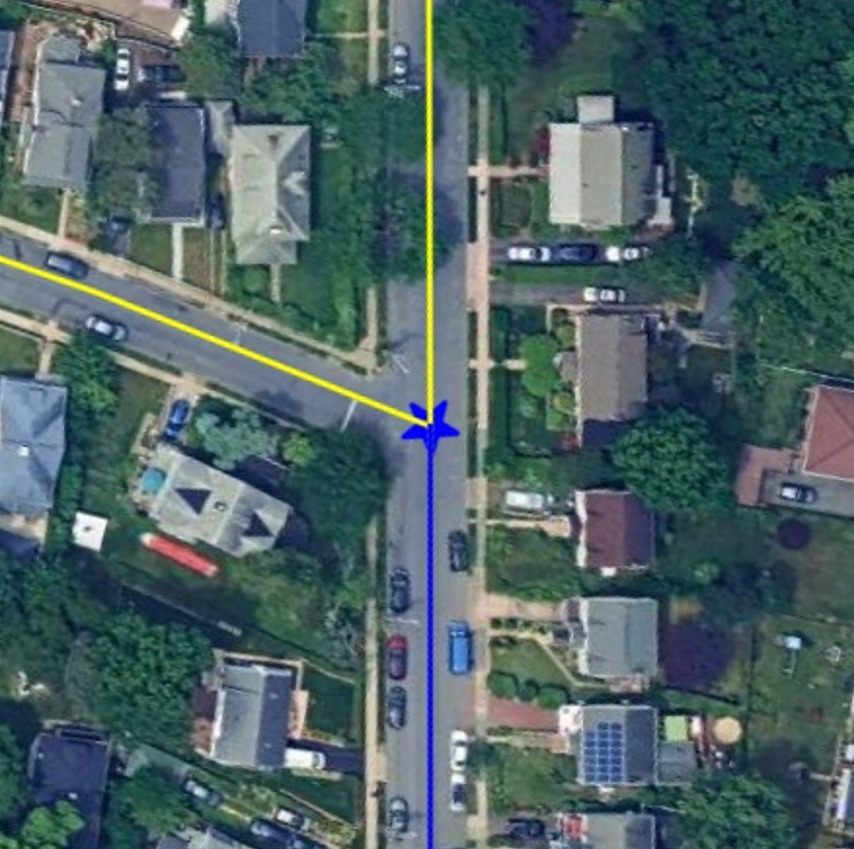}
        \caption{Y-shaped intersection}
    \end{subfigure}\hfill
    \begin{subfigure}[t]{0.23\textwidth}
        \centering
        \includegraphics[width=\linewidth]{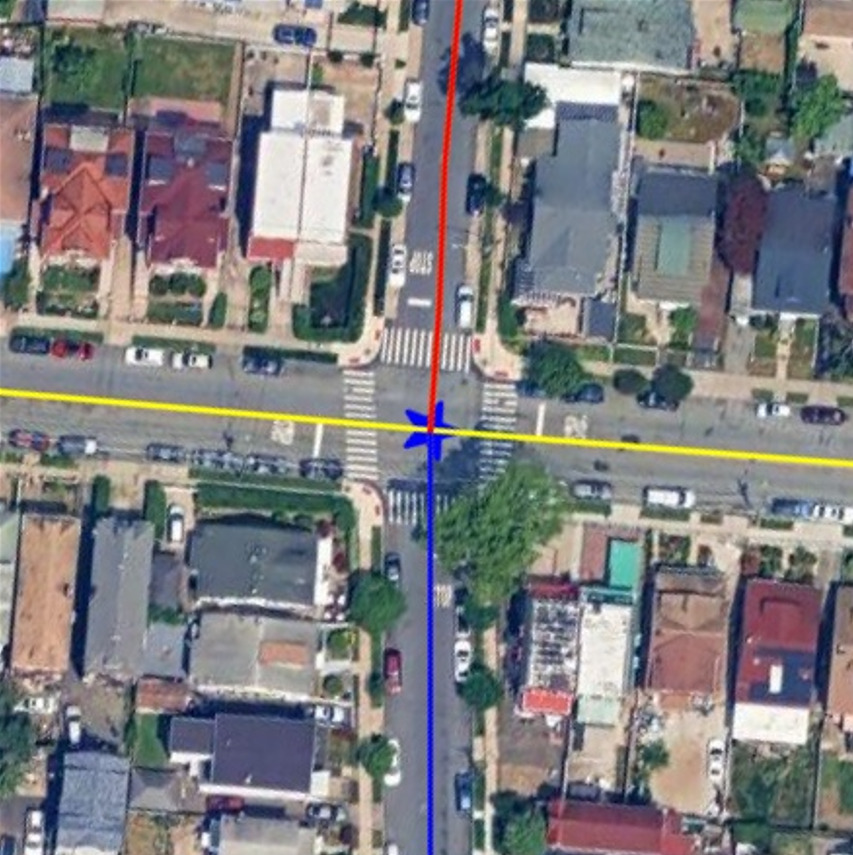}
        \caption{Crossroad}
    \end{subfigure}

    \vspace{0.6em}

    \begin{subfigure}[t]{0.23\textwidth}
        \centering
        \includegraphics[width=\linewidth]{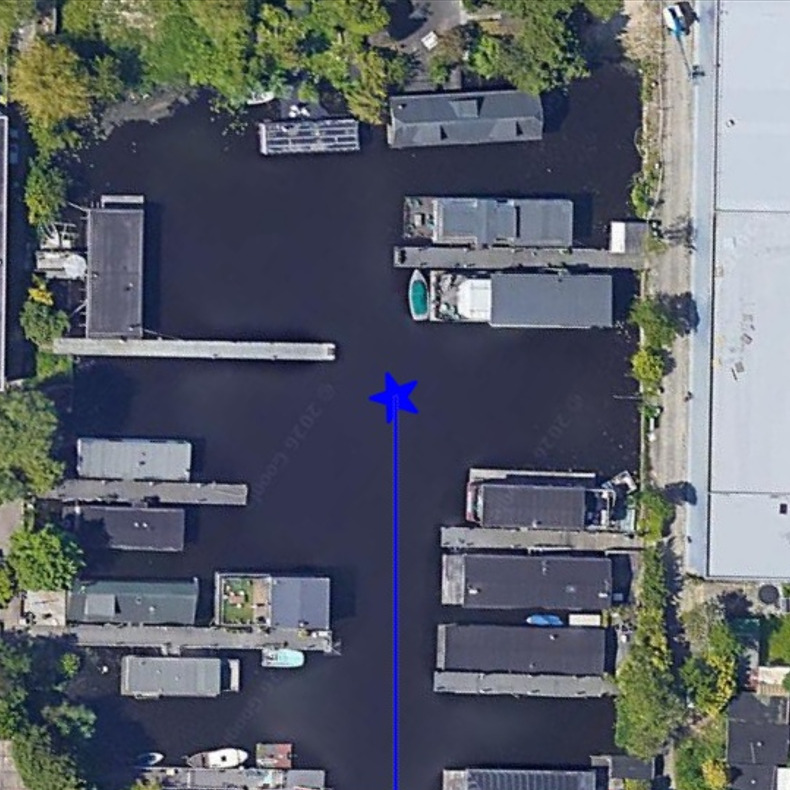}
        \caption{River dead end}
    \end{subfigure}\hfill
    \begin{subfigure}[t]{0.23\textwidth}
        \centering
        \includegraphics[width=\linewidth]{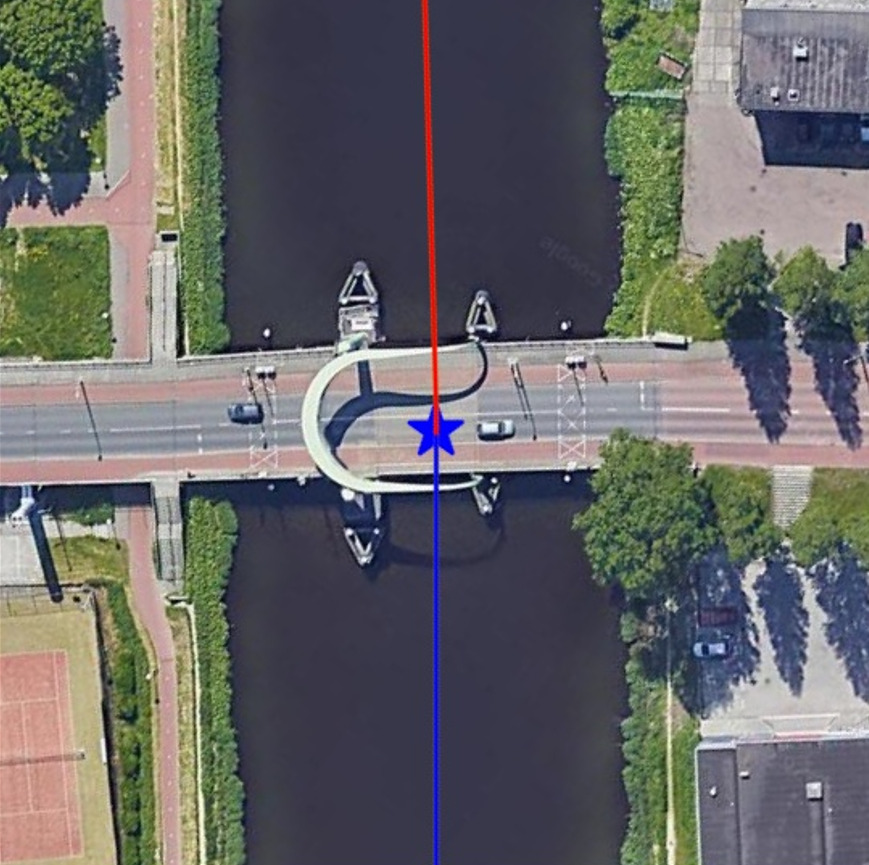}
        \caption{Bridge crossing}
    \end{subfigure}\hfill
    \begin{subfigure}[t]{0.23\textwidth}
        \centering
        \includegraphics[width=\linewidth]{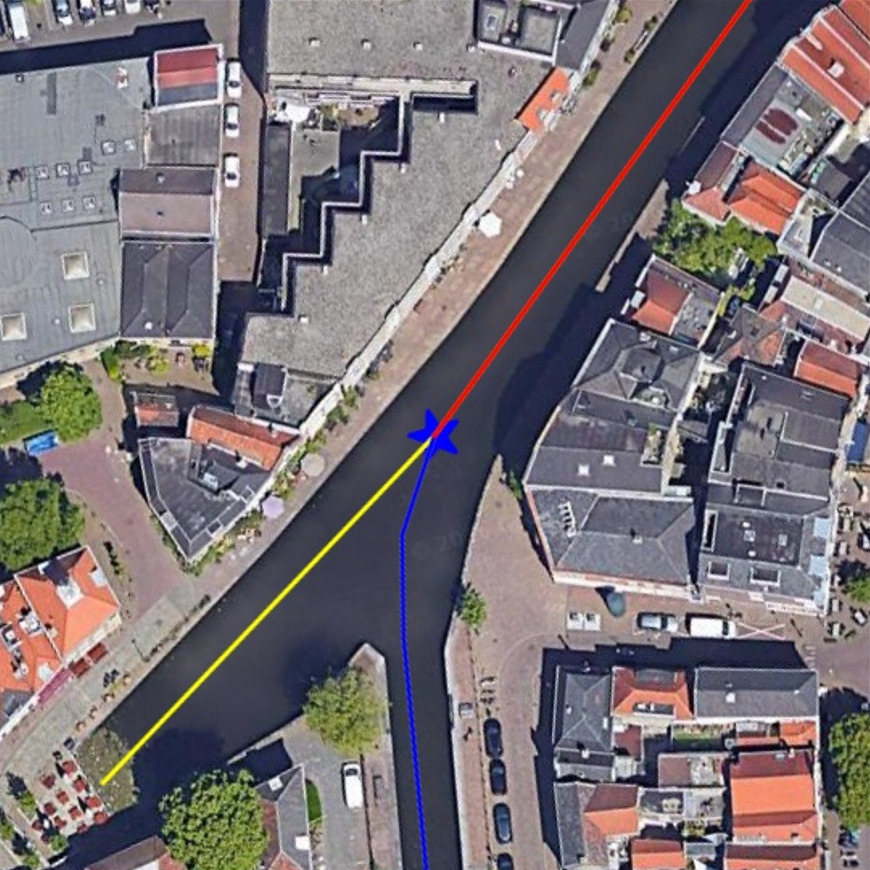}
        \caption{Y-shaped confluence}
    \end{subfigure}\hfill
    \begin{subfigure}[t]{0.23\textwidth}
        \centering
        \includegraphics[width=\linewidth]{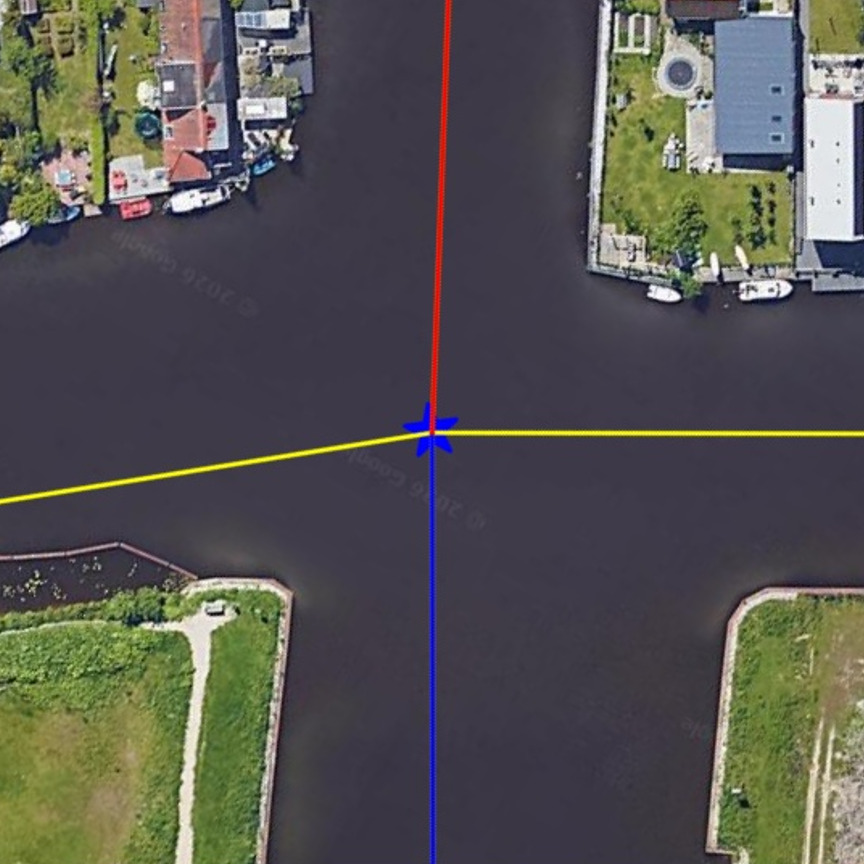}
        \caption{X-shaped confluence}
    \end{subfigure}

    \caption{Representative node types in the road and waterway graphs.}
    \label{fig:road_node_types}
    \vspace{-10pt}
\end{figure*}

Route episodes are constructed from OSM roads, waterways, and their combinations. All line annotations are converted into a metric graph by merging fragmented ways, splitting ways at true intersections, and assigning node and edge attributes from the local topology and OSM semantic labels. Representative node types are shown in Figure~\ref{fig:road_node_types}, and the mapping from local graph structure to node category is summarized in Table~\ref{tab:road_node_mapping}. The resulting graph supports both route-following instructions and counting-based instructions, such as turning at an ordinal intersection or stopping after a specified bridge.

\begin{table*}[t]
\centering
\small
\setlength{\tabcolsep}{6pt}
\caption{Mapping from local graph structure to node categories in the road and waterway graphs.}
\vspace{5pt}
\label{tab:road_node_mapping}
\begin{tabular}{llll}
\toprule
\textbf{Way Type} & \textbf{Degree} & \textbf{Local Geometry} & \textbf{Node Category} \\
\midrule
Highway  & 1       & --             & Road dead end \\
Highway  & 2       & --             & Waypoint \\
Highway  & 3       & T-like         & T-shaped intersection \\
Highway  & 3       & Y-like         & Y-shaped intersection \\
Highway  & 4       & X-like         & Crossroad \\
Waterway & 1       & --             & River dead end \\
Waterway & 2       & --             & Bridge \\
Waterway & 3       & T-like         & T-shaped confluence \\
Waterway & 3       & Y-like         & Y-shaped confluence \\
Waterway & 4       & X-like         & X-shaped confluence \\
\bottomrule
\end{tabular}
\end{table*}

Valid route trajectories are sampled as paths over this graph. We retain trajectories whose key nodes and route segments are visible in local observations, whose nearby topology does not introduce ambiguous counting cues, and whose geometry is sufficiently smooth to be followed by the discrete action space. These checks remove cases where the satellite view cannot reliably support the intended instruction, such as visually unresolved minor structures, weak split-merge junctions, or highly tortuous segments. The retained paths are serialized into route instructions using edge types, node categories, turn directions, and ordinal positions along the trajectory.

\subsection{Episode Quality Review}
\label{app:quality_review}

\begin{figure*}[t]
    \centering
    \includegraphics[width=\textwidth]{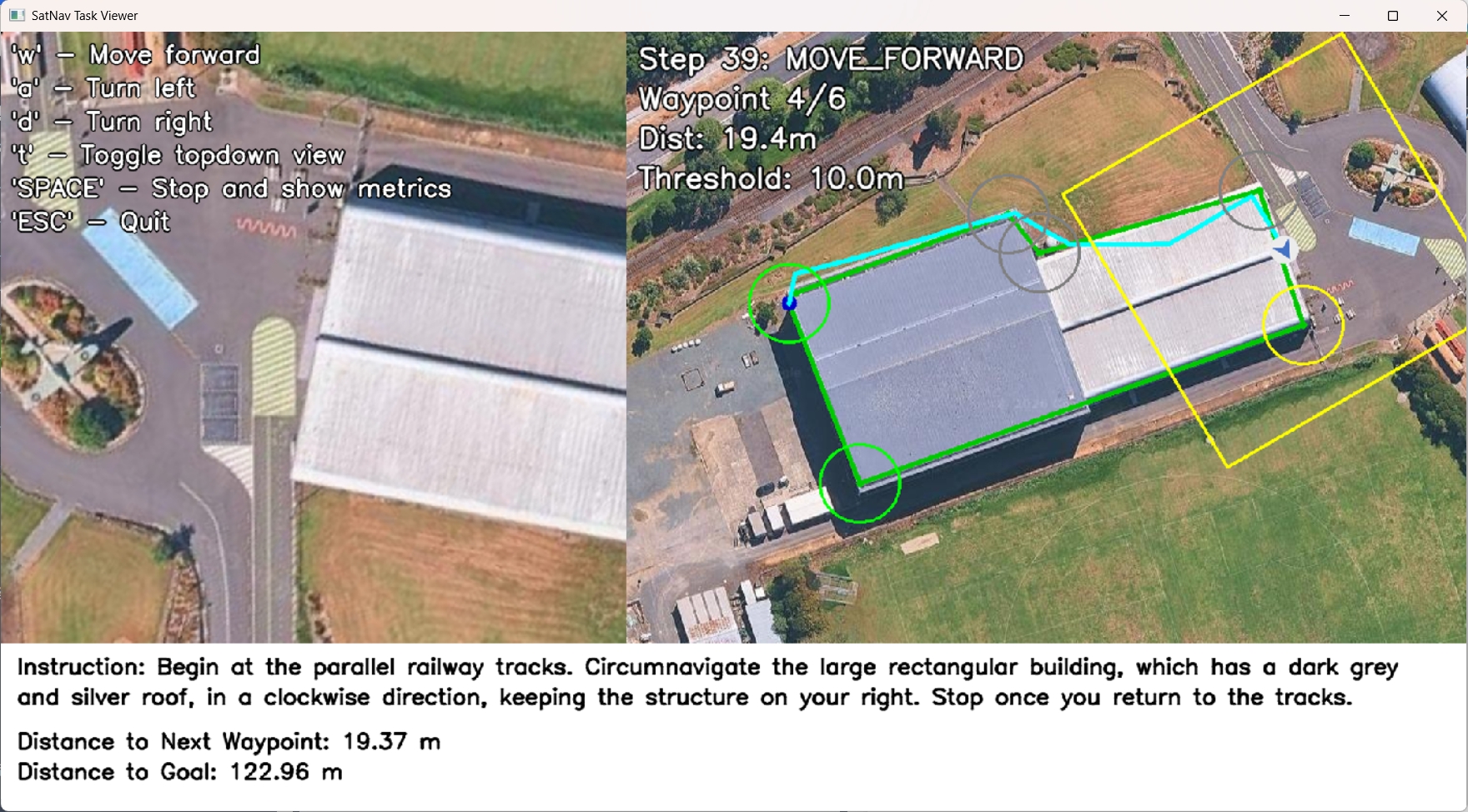}
    \caption{SatNav episode-quality review interface}
    \label{fig:episode_quality_review}
    \vspace{-8pt}
\end{figure*}

To assess instruction-trajectory alignment after the automatic consistency checks described in Section~\ref{sec:episode_generation}, we manually audit 3,000 candidate episodes, randomly sampling 1,000 from each task family before manual filtering. The sample covers all task subtypes, with the composition reported in Table~\ref{tab:episode_quality_audit}. Reviewers inspect each episode in the SatNav task viewer (shown in Figure~\ref{fig:episode_quality_review}), replaying the reference trajectory and checking the task family, subtype, key decision points, spatial and counting cues, and stopping condition. Episodes with inconsistent instruction-trajectory alignment are rejected and removed before release.

\begin{table*}[t]
\centering
\caption{Manual quality audit of 3,000 candidate episodes. Counts and rejection rates are reported by task family and subtype. All rejected episodes are removed before release.}
\label{tab:episode_quality_audit}
\small
\setlength{\tabcolsep}{5pt}
\renewcommand{\arraystretch}{1.15}
\begin{tabular}{llrrrr}
\toprule
Task & Subtype & Audited & Accepted & Rejected & Rejection (\%) \\
\midrule
Boundary & Partial-arc & 155 & 144 & 11 & 7.1 \\
 & Full-loop & 686 & 650 & 36 & 5.2 \\
 & Extended-loop & 159 & 147 & 12 & 7.5 \\
 & \textbf{Total} & \textbf{1,000} & \textbf{941} & \textbf{59} & \textbf{5.9} \\
\midrule
Landmark & One-turn & 813 & 741 & 72 & 8.9 \\
 & Two-turn & 187 & 170 & 17 & 9.1 \\
 & \textbf{Total} & \textbf{1,000} & \textbf{911} & \textbf{89} & \textbf{8.9} \\
\midrule
Route & Road-only & 775 & 769 & 6 & 0.8 \\
 & Waterway-only & 128 & 118 & 10 & 7.8 \\
 & Hybrid & 97 & 90 & 7 & 7.2 \\
 & \textbf{Total} & \textbf{1,000} & \textbf{977} & \textbf{23} & \textbf{2.3} \\
\midrule
\textbf{Overall} & & \textbf{3,000} & \textbf{2,829} & \textbf{171} & \textbf{5.7} \\
\bottomrule
\end{tabular}
\end{table*}

Table~\ref{tab:episode_quality_audit} summarizes the audit results. Overall, 2,829 of the 3,000 sampled candidates are accepted (94.3\%). Rejection rates are 5.9\% for Boundary, 8.9\% for Landmark, and 2.3\% for Route. These statistics describe candidate quality before manual filtering.

\section{Instruction Generation Prompts}
\label{app:instruction_generation_prompts}

SatNav uses four reusable prompt templates for visual grounding and instruction construction. These templates separate cue perception from language rewriting: VLM prompts describe or localize visible cues, while LLM prompts rewrite validated trajectory records without changing navigation semantics.

\paragraph{Visual cue description prompt.}
This prompt is used when a satellite crop contains a marked target region. The VLM is instructed to treat the overlay as an annotation, classify the enclosed visual entity, and return multi-level descriptions.

\begin{promptbox}
\footnotesize
\textbf{Role:}\\
You are a rigorous geographic analysis expert specializing in
remote sensing and aerial image interpretation.

\textbf{Input:}\\
You will be given a satellite image containing a circular red line.
Important: the red line is only an annotation and is NOT part of the real scene.

\textbf{Objective:}\\
Analyze the area enclosed by the red line. Determine whether it is a single,
distinct visual object category and then describe the enclosed region at three
levels of detail.

\textbf{Step 1. Category classification:}\\
Decide whether the enclosed area is exactly one instance of one of the
following categories:
\begin{itemize}[leftmargin=1.5em, itemsep=1pt, topsep=2pt]
    \item[-] \texttt{Building}, \texttt{Water Body}, \texttt{Farmland}, \texttt{Parking Lot}, \texttt{Sports Field}, \texttt{Islet}, \texttt{Other}
\end{itemize}

\textbf{Step 2. Multi-level description:}\\
Provide three descriptions:
\begin{itemize}[leftmargin=1.5em, itemsep=1pt, topsep=2pt]
\item[-] \texttt{telegraphic: ultra-concise; object type + most salient visual trait.}
\item[-] \texttt{natural: moderate detail; shape, color, texture, and nearby context.}
\item[-] \texttt{academic: formal technical description; morphology, spectral appearance, boundary clarity, and spatial context.}
\end{itemize}
\textbf{Critical constraints:}
\begin{itemize}[leftmargin=1.5em, itemsep=1pt, topsep=2pt]
\item[-] Never mention the red line.
\item[-] Treat all overlays, masks, markers, and grids as annotations, not objects.
\item[-] Keep the descriptions grounded in visible evidence only.
\end{itemize}
\end{promptbox}

\paragraph{Landmark discovery and coarse localization prompt.}
This prompt searches a broad grid image for distinctive navigation anchors. The prompt uses a closed-category policy and an explicit re-audit step to avoid repeated, weak, or ambiguous landmarks.

\begin{promptbox}
\footnotesize
\textbf{Role:}\\
You are an expert in drone visual navigation.

\textbf{Input:}\\
You are given an aerial $8\times8$ grid image. Each grid cell has a visible \texttt{(Row, Col)} coordinate in yellow text. Each grid cell is approximately $40\,\mathrm{m}\times40\,\mathrm{m}$.

\textbf{Objective:}\\
Extract prominent and reliable navigation anchors from the image.

\textbf{Closed-category landmark policy:}\\
Select \textbf{ONLY} from the following categories:

\begin{itemize}[leftmargin=1.8em, itemsep=1pt, topsep=2pt]
    \item[-] \texttt{High-Contrast Roofs}, \texttt{Unique Vegetation}, \texttt{Specialized Water Features}, \\ \texttt{Sport Courts}, \texttt{Featured Facilities on Large Backgrounds}
\end{itemize}

\textbf{Strict selection rules:}
\begin{itemize}[leftmargin=1.5em, itemsep=1pt, topsep=2pt]
    \item A landmark must be unique in the entire image, visually salient and easy to recognize, and roughly between $10\,\mathrm{m}\times10\,\mathrm{m}$ and $60\,\mathrm{m}\times60\,\mathrm{m}$ in size.
    \item Repeated roofs, ordinary vegetation, large lakes, and ambiguous shadows do not qualify.
\end{itemize}

\textbf{Workflow:}
\begin{enumerate}[leftmargin=1.8em, itemsep=1pt, topsep=2pt]
    \item Identify up to 10 candidate landmarks.
    \item Re-audit every candidate and reject anything too small, too large, low-contrast, duplicated, or visually ambiguous.
    \item Return each surviving landmark with its grid coordinate and a concise descriptive phrase.
\end{enumerate}
\end{promptbox}

\paragraph{Landmark verification and fine localization prompt.}
This prompt verifies a proposed landmark in a local $3\times3$ crop. It returns exactly one localization token or \texttt{null}; no explanation is allowed.

\begin{promptbox}
\footnotesize
\textbf{Role:}\\
You are a vision navigation expert.

\textbf{Input:}\\
You are given a local satellite crop with a $3\times3$ grid overlay.\\
The target landmark is described as:\\
\textit{``Bright red rectangular roof with strong contrast against gray surroundings.''}

\textbf{Task:}\\
Locate the target landmark in the image.

If the landmark is found, return \textbf{EXACTLY ONE} token from:
\begin{center}
{\ttfamily
UL \quad UM \quad UR\\
ML \quad C  \quad MR\\
LL \quad LM \quad LR
}
\end{center}
If the landmark is absent, too small, low-contrast, or visually inconspicuous, return: \texttt{null}.

\textbf{Definitions:}
\begin{itemize}[leftmargin=1.5em, itemsep=1pt, topsep=2pt]
    \item \texttt{UL} = upper-left, \texttt{UM} = upper-middle, \texttt{UR} = upper-right.
    \item \texttt{ML} = middle-left, \texttt{C} = center, \texttt{MR} = middle-right.
    \item \texttt{LL} = lower-left, \texttt{LM} = lower-middle, \texttt{LR} = lower-right.
\end{itemize}

\textbf{Critical rules:}
\begin{itemize}[leftmargin=1.5em, itemsep=1pt, topsep=2pt]
    \item Reject false detections aggressively.
    \item Output must be exactly one word. No explanation, no punctuation, no markdown.
    \item If the supposed landmark could actually be a shadow, texture patch, or weak cue, return \texttt{null}.
\end{itemize}

\end{promptbox}

\paragraph{Instruction rewriting prompt.}
After trajectory facts are validated, the rewriting prompt modifies only the surface form of the instruction. The implementation uses three output fields, \texttt{telegraphic}, \texttt{natural}, and \texttt{academic}, which correspond to compact, fluent, and descriptive instruction styles.

\begin{promptbox}
\footnotesize
\textbf{Task:}\\
Rewrite the validated navigation record below into three styles. Preserve all navigational semantics exactly; only the phrasing may change.

\textbf{Shared constraints:}
\begin{itemize}[leftmargin=1.5em, itemsep=1pt, topsep=2pt]
    \item[-] Preserve all route facts exactly, and do not add unsupported landmarks, scene details, or explanations.
    \item[-] Keep direction words, route-type terminology, and the original action order unchanged.
    \item[-] Maintain the original granularity of navigation steps while increasing syntactic diversity across outputs.
\end{itemize}

\textbf{Structured trajectory facts:}

\begin{itemize}[leftmargin=1.5em, itemsep=1pt, topsep=2pt]
    \item \texttt{trajectory type}: Hybrid
    \item step 1: start from the T-junction, move forward along the road
    \item step 2: turn right at the first crossroad, continue forward along the riverbank
    \item step 3: turn left at the second T-junction, continue forward along the road
    \item step 4: stop at the third intersection
\end{itemize}

\textbf{Style targets:}
\begin{itemize}[leftmargin=1.5em, itemsep=1pt, topsep=2pt]
    \item \texttt{telegraphic}: concise, action-first, compact mission phrasing
    \item \texttt{natural}: fluent and human-like, but still exact
    \item \texttt{academic}: more descriptive and formally structured
\end{itemize}
\end{promptbox}

\section{SwiftVLN Framework Details}
\label{app:swiftvln_details}

\subsection{Overall Pipeline}
\label{app:swiftvln_pipeline}
To formalize SwiftVLN's data flow, we distinguish physical simulator steps from model inferences. Following UniNaVid~\citep{zhang2024uni} and StreamVLN~\citep{wei2025streamvln}, our framework employs an action sequence generation mechanism rather than predicting a single action. Thus, we define the time index $t$ strictly as a \textit{model-query round}, i.e., the discrete moment when the backbone processes inputs to generate an action sequence.

At any given query round $t$, the model takes four inputs: the textual instruction $\mathcal{I}$, the current visual observation $o_t$, the short-term dialogue memory $S_t$ (a sliding window of recent image-action turns), and the long-term memory $L_t$ (pre-window historical observations processed by a memory module). The prompt construction module $\Phi$ assembles these into a unified sequence $C_t$:
$$C_t = \Phi(\mathcal{I}, o_t, S_t, L_t)\ .$$
The visual-language backbone $f_\theta$ processes this context to sequentially decode a textual response $Y_t$:
$$Y_t = f_\theta(C_t)\ .$$
This response is parsed into an action chunk $A_t = [a_{t,1}, \ldots, a_{t,k}]$, comprising a sequence of low-level navigation actions. During evaluation, these actions are executed sequentially in the environment, triggering the next model query (round $t+1$) only when the queue is depleted.

Following the generation of $Y_t$, the memories are dynamically managed. The short-term memory is updated by appending the newly generated turn (i.e., the current visual observation $o_t$ and the assistant response $Y_t$) to the multi-turn dialogue window:
$$S_{t+1} = \text{Slide}(S_t \oplus [o_t, Y_t])\ .$$
Here, the $\text{Slide}(\cdot)$ operation enforces a maximum window capacity of $N_w$ turns. Once this capacity is reached, the window slides forward by removing the oldest turns while retaining a predefined number of overlapping turns $N_o$. This ensures that recent context is preserved, maintaining local continuity across successive windows.

Conversely, the long-term memory $L_t$ is updated only when the window slides. It is reconstructed from $\mathcal{H}_t$, which denotes the complete trajectory of historical observations up to the new window boundary, and %
then processed by a variant-specific memory module $\mathcal{M}$:
$$L_{t+1} = \begin{cases}
\mathcal{M}(\mathcal{H}_t), & \text{if window slides} \\
L_t, & \text{otherwise}
\end{cases}\ .$$

In our implementation, the visual observation $o_t$ is a $448 \times 448$ cropped satellite image, and the visual-language backbone $f_\theta$ is instantiated as Qwen2.5-VL-3B-Instruct. Following StreamVLN, the decoded response $Y_t$ is represented as a sequence of direction symbols drawn from $\{\uparrow, \leftarrow, \rightarrow, \texttt{STOP}\}$, from which the action chunk $A_t$ is parsed. At each model-query round $t$, the model generates an action chunk of fixed length $k=4$. The short-term memory capacity is set to $N_w=8$. For the SwiftVLN baseline, we set the overlap size to $N_o=0$, following the original StreamVLN setting.

\subsection{Prompt Construction}
\begin{promptbox}
\textbf{System}: You are an autonomous navigation assistant. Your task is to \textcolor{PromptInstColor}{Take off from the group of parked vehicles. Fly counter-clockwise around the large rectangular maintenance shed, keeping the structure to your left. Stop once the parked vehicles are visible again}. Based on your observations, output a sequence of actions using: $\uparrow$ (forward 10m), $\leftarrow$ (turn left), $\rightarrow$ (turn right), or STOP (when goal is reached). Output actions directly without explanation. \textcolor{PromptMemColor}{These are your historical observations: $\langle$history\_memory$\rangle$}.

\tcblower
\textcolor{PromptShortColor}{%
\textbf{User}: Below you is $\langle$image$\rangle$.\\
\textbf{Assistant}: $\uparrow\uparrow\rightarrow\uparrow$\\[0.5em] %
\textbf{User}: In your sight is $\langle$image$\rangle$.\\
\textbf{Assistant}: $\uparrow\uparrow\leftarrow\leftarrow$\\[0.5em]
\textbf{User}: The current view covers $\langle$image$\rangle$.\\
\textbf{Assistant}:
}
\end{promptbox}

An example prompt constructed by the prompt construction module $\Phi$ is shown above. The \textcolor{PromptInstColor}{blue text} represents the task instruction $\mathcal{I}$, while the \textcolor{PromptMemColor}{yellow segment} corresponds to the long-term memory prompt derived from $L_t$. The \textcolor{PromptShortColor}{green block} contains the active short-term dialogue context, consisting of two completed image-action turns from $S_t$ followed by the current user turn associated with $o_t$. Together, these components are assembled into the context $C_t$ for decoding the next response $Y_t$.

\subsection{Memory Design Details}
\label{app:swiftvln_memory_details}

\textbf{History Frame Sampling} selects a subset of frames from the historical frame set $\mathcal{H}_t$. We support three strategies: uniform, random, and temporal-biased sampling. For the temporal-biased strategy, we set the sampling exponent to 2, so that recent frames are sampled more densely than distant ones. By default, we retain 8 historical frames. Each retained frame is encoded by the shared vision encoder and compressed by average pooling into 64 tokens, resulting in 512 history tokens that are inserted into $\langle \text{history\_memory} \rangle$. Unless otherwise specified, the subsequent memory variants use the same default history source, namely 8 uniformly sampled frames from $\mathcal{H}_t$.

\textbf{Initial Frame Prompting} adds \textcolor{PromptMemColor}{``This is your initial observation at the starting point of this journey: $\langle \text{initial\_image} \rangle$''} to the system prompt. The first image is encoded into 256 visual tokens and inserted into $\langle \text{initial\_image} \rangle$, while $\langle \text{history\_memory} \rangle$ uses the default uniform sampling.

\textbf{Pose Encoding} augments each incoming image with a 4D relative pose vector $\mathbf{p}_i=[\tanh(\Delta^{\text{fwd}}_i/100),\tanh(\Delta^{\text{right}}_i/100),\sin(\Delta^\theta_i),\cos(\Delta^\theta_i)]$,
where the pose is defined relative to the episode start. The translation terms are scaled by 100 before the \texttt{tanh} operation to keep the positional values in a bounded numerical range.
A two-layer MLP with hidden size 256 projects $\mathbf{p}_i$ to FiLM~\citep{perez2018film} parameters $[\gamma_i,\beta_i]=\mathrm{MLP}(\mathbf{p}_i)$, which modulate the visual tokens of image $i$ as $\mathbf{X}'_i=\mathbf{X}_i \odot (1+\gamma_i)+\beta_i$. Here, $\gamma_i$ and $\beta_i$ are broadcast across all tokens of the image, and the final linear layer is zero-initialized so that the module starts as a no-op.

\textbf{Global Token Clustering (GTC)} compresses long-term history by clustering the merged visual tokens from historical observations into a fixed-size memory block. Instead of using the default 8 uniformly sampled frames, GTC uses all previous image-action observations in $\mathcal{H}_t$ before the current window start. After visual encoding, all resulting history tokens are concatenated and clustered jointly with cosine-similarity soft k-means into 512 tokens. In our implementation, GTC uses uniform centroid initialization, a temperature of $0.1$, and one clustering iteration. The resulting clustered tokens are then inserted into $\langle \text{history\_memory} \rangle$.

\textbf{Segment Token Clustering (STC)} compresses long-term history by clustering historical tokens in a segment-wise manner. Similar to GTC, STC uses all previous image-action observations in $\mathcal{H}_t$. After visual encoding, the historical frames are divided into 8 temporal segments, and each segment is clustered independently with cosine-similarity soft k-means. The total token budget is 512, which is distributed across segments, and the resulting segment-level tokens are concatenated from early to late before being inserted into $\langle \text{history\_memory} \rangle$. Compared with GTC, this preserves a coarse chronological structure in the memory representation.

\begin{figure*}[t]
\centering

\begin{subfigure}[t]{0.45\textwidth}
\centering
\includegraphics[width=0.48\linewidth]{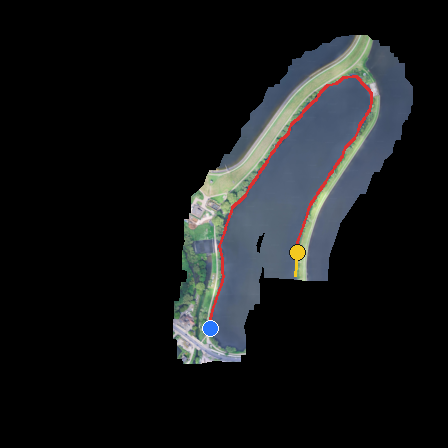}
\hfill
\includegraphics[width=0.48\linewidth]{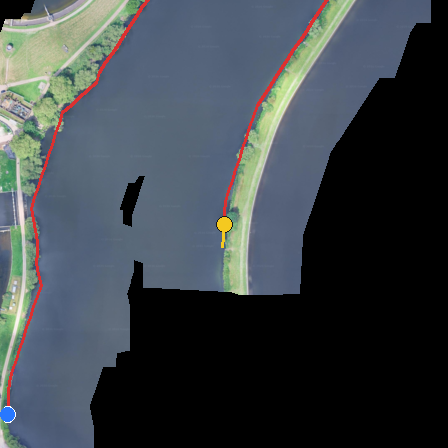}
\caption{\small London-2\_Boundary\_ID-1279}
\label{fig:map_memory_london}
\end{subfigure}
\hspace{0.06\textwidth}
\begin{subfigure}[t]{0.45\textwidth}
\centering
\includegraphics[width=0.48\linewidth]{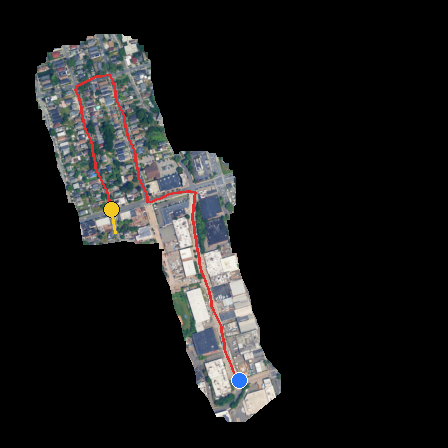}
\hfill
\includegraphics[width=0.48\linewidth]{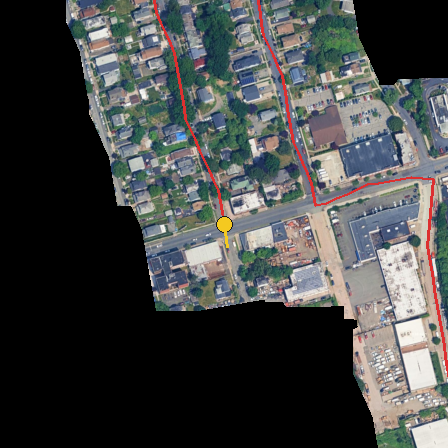}
\caption{\small NewYork-1\_Route\_ID-999}
\label{fig:map_memory_newyork}
\end{subfigure}

\caption{\small Examples of map memory. Blue indicates the start point, yellow indicates the current position and heading, and red indicates the trajectory. In each pair, the left is the global map and the right is the local map.}
\label{fig:map_memory_examples}
\end{figure*}

\textbf{Map Memory} replaces history frames with two top-down north-up maps: a start-centered global map and a current-centered local map. In the system prompt, the long-term memory description is replaced with \textcolor{PromptMemColor}{``These are your explored map memories: $\langle \text{history\_memory} \rangle$''}. Each map is encoded by the shared vision encoder into 256 visual tokens, yielding 512 tokens in total. The maps are constructed from the orthophoto and accumulated trajectory with unexplored regions masked out, using 1000\,m and 400\,m crops for the global and local maps. Examples are shown in Figure~\ref{fig:map_memory_examples}.

\subsection{Training Details}
\label{app:swiftvln_training_details}

We train the SwiftVLN baseline and its variants on 8$\times$ NVIDIA H100 80GB GPUs. The specific hyperparameters for the SwiftVLN baseline are summarized in Table~\ref{tab:lvlm_training_details}. Note that while the choice of different {memory modules $\mathcal{M}$} may lead to variations in training cost (GPU hours), the core optimization hyperparameters remain consistent across all configurations to ensure a fair comparison.

\subsection{Satellite-to-UAV Adapter Details}
\label{app:swiftvln_sim2real_details}

\begin{figure*}[t]
\centering
\includegraphics[width=\textwidth]{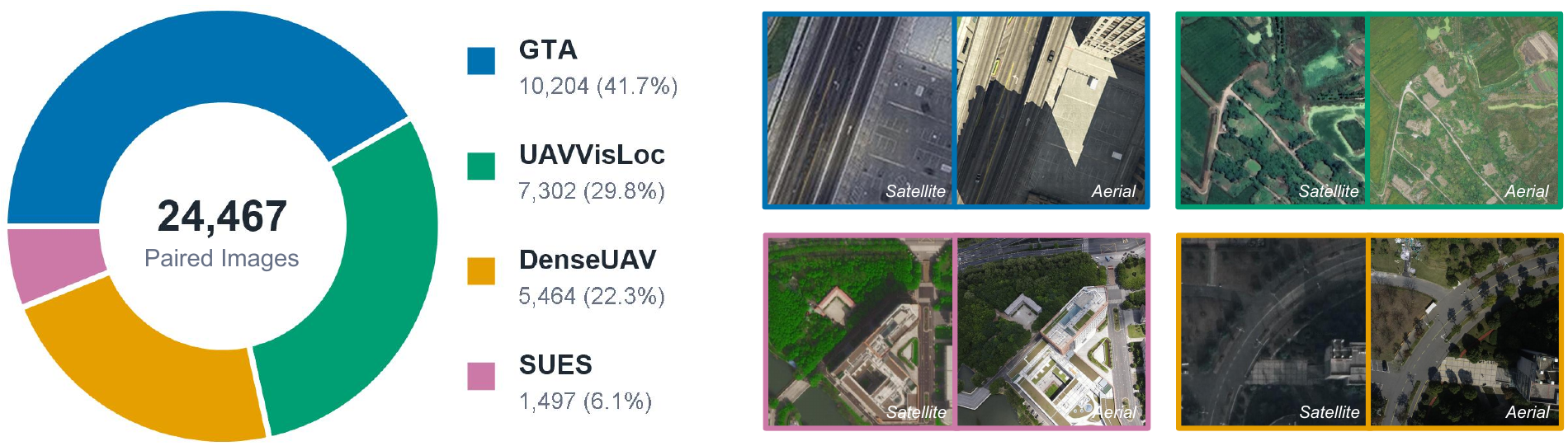}
\caption{\small Data distribution and examples of the paired UAV-satellite images.}
\label{fig:sim2real_data}
\end{figure*}

We train our satellite-to-UAV adapter on 24,467 paired UAV-satellite images compiled from DenseUAV~\citep{dai2023vision}, GTA-UAV~\citep{ji2025game4loc}, SUES~\citep{zhu2023sues}, and UAV-VisLoc~\citep{xu2024uav}, with 21,216 training pairs and 3,251 validation pairs. During pair construction, each satellite image is aligned with its paired UAV image in both viewing extent and orientation. Data distribution and examples are shown in Figure~\ref{fig:sim2real_data}.

For a paired sample $(x_i^u, x_i^s)$, a frozen Qwen2.5-VL visual tower $f_{\mathrm{teach}}$ first extracts token sequences $U_i=f_{\mathrm{teach}}(x_i^u)$ and $S_i=f_{\mathrm{teach}}(x_i^s)$. Only the UAV branch is passed through a trainable token-level Transformer adapter $A_{\phi}$, producing $\tilde{U}_i=A_{\phi}(U_i)$, while the satellite branch remains unchanged as a stable target. We implement $A_{\phi}$ as a 2-layer token-level Transformer with 8 attention heads, an MLP expansion ratio of 4.0, and dropout 0.0. We then apply masked mean pooling to $\tilde{U}_i$ and $S_i$ to obtain image-level global features $\bar{u}_i$ and $\bar{s}_i$.

The two losses are computed at different levels. The cosine loss $L_{\text{cos}}$ is applied directly to the normalized pooled features, encouraging the adapted UAV global representation to stay close to its paired satellite representation. In contrast, the bidirectional contrastive loss $L_{\text{contrast}}$ is computed in a shared projection space: a shared projection head $P_{\psi}$ first maps $\bar{u}_i$ and $\bar{s}_i$ to projected embeddings, which are then L2-normalized to produce $z_i^u$ and $z_i^s$ for contrastive learning. The final training objective combines the two losses as $L = L_{\text{contrast}} + L_{\text{cos}}$.

We optimize only the adapter $A_{\phi}$ and projection head $P_{\psi}$ with AdamW. The default training setup uses a learning rate of $1\times10^{-4}$, a weight decay of 0.01, a warmup ratio of 0.05, and 10 epochs. Validation uses cross-view retrieval metrics, including UAV-to-satellite and satellite-to-UAV Recall@1/5/10, together with the mean cosine similarity of matched pairs. The best checkpoint is selected as the one with the highest validation UAV-to-satellite Recall@1.

\section{Baseline Adaptation and Training Details}
\label{app:baseline}

\subsection{Baseline Adaptation}
\label{app:baseline_adaptation}

To ensure a fair comparison on SatNav, we adapt all baselines to a unified input-output formulation for both training and evaluation. The shared action space strictly consists of \texttt{forward} (10 meters), \texttt{left} ($15^\circ$), \texttt{right} ($15^\circ$), and \texttt{stop}. At each decision step, models process the task instruction alongside visual observations (either a single $448\times448$ cropped satellite image or a sequence) to predict navigation actions.
Unless otherwise specified, all LVLM baselines are trained offline via teacher-forcing using their original causal language modeling objectives. Table~\ref{tab:baseline_adaptation} summarizes the specific architectural adaptations for each baseline.

\begin{table*}[t]
\centering
\caption{Summary of baseline adaptations for SatNav}
\label{tab:baseline_adaptation}
\resizebox{\linewidth}{!}{
\begin{tabular}{lccc}
\toprule
\textbf{Method} & \textbf{Visual Input per Step} & \textbf{Action Output Horizon} & \textbf{Action Format} \\
\midrule
Seq2Seq \& CMA & 1 current & 1 step & Categorical Logits \\
NaVILA & 1 current + 7 history & 1 step & Single-word text action \\
UniNaVid & 1 current + all history & 4 steps & Single-word text action \\
StreamVLN & current window + 8 history & 4 steps & Symbolic (e.g., $\leftarrow, \uparrow$) \\
OpenFly-Agent & 1 current + 2 history + 16 past actions & 1 step & Single-word text action \\
\bottomrule
\end{tabular}
}
\end{table*}

\textbf{Seq2Seq \& CMA}. We retain their core VLN-CE architectures, including 50-dimensional GloVe embeddings and \texttt{torchvision} ImageNet-pretrained ResNet visual encoders using solely RGB inputs. We discard the original online recollection and auxiliary progress-monitor losses. Instead, both models are trained with an empirically weighted cross-entropy loss over the unified action space. Specifically, for Seq2Seq, the class weights for \texttt{forward}, \texttt{left}, \texttt{right}, and \texttt{stop} are manually set to 1.0, 1.5, 1.5, and 2.0, respectively. For CMA, to explicitly suppress early-stop collapse, we aggressively downscale the \texttt{stop} weight to 0.3 while maintaining identical directional weights.

\textbf{NaVILA}. While preserving the original prompt scaffold, we rewrite the final instruction to align with SatNav. The visual input comprises a fixed 8-frame sequence: the current observation and 7 frames sampled from the trajectory history. To replace the original free-form natural-language descriptions (e.g., ``The next action is turn left 15 degrees''), the model is softly constrained, via prompt wording and single-word supervision, to output a single valid action word. During inference, the predicted action is parsed directly from the greedy-decoded text.

\textbf{UniNaVid}. We adapt its original multimodal execution to non-overlapping four-step windows. Instead of per-step queries, the system interacts with the model once per window. It processes the instruction and accumulated observation history to autoregressively generate a four-step text sequence explicitly formatted with step numbers. During evaluation, the model executes this action queue sequentially; a new observation is acquired and a decision query is formulated only when the queue is exhausted or the episode terminates early.

\textbf{StreamVLN}. To adapt the observation domain, we replace standard inputs with cropped satellite RGBs while preserving the multi-frame formulation. The model constructs a memory context using 8 frames uniformly sampled from the prior trajectory history, which is then combined with current-window observations. It autoregressively outputs four actions in symbolic format.

\textbf{OpenFly-Agent}. The autoregressive training pipeline remains largely unchanged, with adaptations focusing on the data interface and training target. Each step receives three frames (the current and two previous) and a text prompt explicitly listing up to 16 past actions. Crucially, to fit the compact SatNav action space, we replace the original tokenized 8-dimensional continuous action vectors with single-word text actions.

\subsection{Baseline Training Details}
\label{app:baseline_training_details}

All baselines are trained on 8$\times$ NVIDIA H100 80GB GPUs. We follow the original public training recipes of each baseline as closely as possible, and adjust the training hyperparameters only when necessary to fit our 8-GPU H100 setup and the unified SatNav setting. Table~\ref{tab:classical_training_details} summarizes the training details of the Seq2Seq and CMA baselines, while Table~\ref{tab:lvlm_training_details} reports those of the LVLM baselines.

\begin{table*}[t]
\centering
\caption{Training configurations for classical baselines on SatNav}
\label{tab:classical_training_details}
\begin{tabular}{lcc}
\toprule
\textbf{Configuration} & \textbf{Seq2Seq} & \textbf{CMA} \\
\midrule
Vision Encoder & \texttt{ResNet50} & \texttt{ResNet50} \\
Instruction Encoding & \texttt{GloVe-50d + GRU} & \texttt{GloVe-50d + BiGRU} \\
Global Batch Size & 64 & 32 \\
Optimizer & \texttt{Adam} & \texttt{Adam} \\
Weight Decay & 0.0 & 0.0 \\
Peak LR & \texttt{3e-4} & \texttt{1e-4} \\
LR Schedule & \texttt{Constant} & \texttt{Constant} \\
Total Epochs & 10 & 10 \\
\bottomrule
\end{tabular}
\end{table*}

\begin{table*}[t]
\centering
\caption{Training configurations for LVLM baselines on SatNav}
\label{tab:lvlm_training_details}
\resizebox{\textwidth}{!}{
\begin{tabular}{>{\raggedright\arraybackslash}m{3.0cm}ccccc}
\toprule
\textbf{Configuration} & \textbf{NaVILA} & \textbf{StreamVLN} & \textbf{UniNaVid} & \textbf{OpenFly-Agent} & \textbf{SwiftVLN} \\
\midrule
Base Model
& \makecell[c]{\texttt{Llama-3-}\\\texttt{VILA1.5-8B}}
& \makecell[c]{\texttt{LLaVA-Video-}\\\texttt{7B-Qwen2}}
& \makecell[c]{\texttt{vicuna-7b-}\\\texttt{v1.5}}
& \makecell[c]{\texttt{openvla-7b-}\\\texttt{prismatic}} 
& \makecell[c]{\texttt{Qwen2.5-VL-}\\\texttt{3B-Instruct}} \\
Vision Encoder & \texttt{SigLIP} & \texttt{SigLIP} & \texttt{EVA-CLIP ViT} & \texttt{DINOv2 + SigLIP} & \texttt{Qwen2.5-VL ViT} \\
LLM Backbone & \texttt{LLaMA-3-8B} & \texttt{Qwen2-7B} & \texttt{Vicuna-7B} & \texttt{LLaMA-2-7B} & \texttt{Qwen2.5-3B} \\
Precision & \texttt{bf16} & \texttt{bf16} & \texttt{bf16} & \texttt{bf16} & \texttt{bf16} \\
Fine-tuning Strategy & \texttt{Full} & \texttt{Full} & \texttt{Frozen Vision} & \texttt{Full} & \texttt{Full} \\
\#GPUs & 8 & 8 & 8 & 8 & 8 \\
Local Batch Size & 4 & 2 & 24 & 12 & 8 \\
Grad Accum. & 1 & 2 & 1 & 1 & 1 \\
Global Batch Size & 32 & 32 & 192 & 96 & 64 \\
Optimizer & \texttt{AdamW} & \texttt{AdamW} & \texttt{AdamW} & \texttt{AdamW} & \texttt{AdamW} \\
Weight Decay & 0.0 & 0.0 & 0.0 & 0.0 & 0.0 \\
Peak LR & \texttt{3e-5} & \texttt{2e-5} & \texttt{1e-5} & \texttt{2e-5} & \texttt{2e-5} \\
LR Scheduler & \texttt{cosine} & \makecell[c]{\texttt{cosine\_with\_}\\\texttt{min\_lr}} & \texttt{cosine} & \texttt{linear} & \makecell[c]{\texttt{cosine\_with\_}\\\texttt{min\_lr}} \\
Warmup Ratio & 0.03 & 0.075 & 0.03 & 0.0 & 0.075 \\
Total Epochs & 0.6 & 1 & 1 & 1 & 1 \\
Total Steps & 60,170 & 6,885 & 7,246 & 39,839 & 3,518 \\
Distributed Training & \texttt{ZeRO-2} & \texttt{ZeRO-2} & \texttt{ZeRO-1} & \texttt{ZeRO-2} & \texttt{ZeRO-2} \\
GPU Hours & 312.1 GPUh & 54.5 GPUh & 60.6 GPUh & 102.9 GPUh & 33.8 GPUh \\
\bottomrule
\end{tabular}
}
\end{table*}

\section{Real-world Demonstration Details}
\label{app:real_world_demo}

\subsection{Real-world Deployment}
\label{app:real_world_deployment}

\begin{figure*}[t]
\centering
\includegraphics[width=0.9\textwidth]{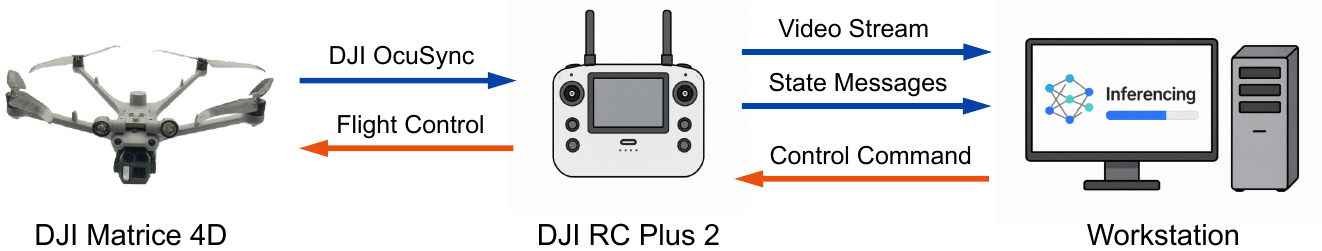}
\caption{\small Real-world operation pipeline.}
\label{fig:real_world_pipeline}
\end{figure*}

We deploy our system on a DJI Matrice 4D for real-world flight tests, as illustrated in Figure~\ref{fig:real_world_pipeline}. During deployment, the onboard image stream and UAV state messages, including longitude, latitude, altitude, and heading, are first transmitted to a DJI RC Plus 2 controller through the DJI OcuSync Enterprise video transmission system. The controller then relays the data to a local workstation over Wi-Fi, using RTMP for video streaming and MQTT for state and control communication. On the workstation, our satellite-to-UAV adapted SwiftVLN baseline, built on Qwen2.5-VL-3B, runs on a single RTX 4090 GPU for online inference.

In practice, the latency between the UAV and the controller is nearly negligible. The main delay comes from forwarding the video stream from the controller to the workstation via RTMP over Wi-Fi, which typically introduces 3-5 seconds of latency due to buffering and network variability. By contrast, model inference on a single RTX 4090 takes only about 250\,ms on average, making the communication pipeline the primary runtime bottleneck in real-world deployment.

\subsection{More Real-world Flight Results}
\label{app:more_real_world_flights}
\begin{figure}[p]
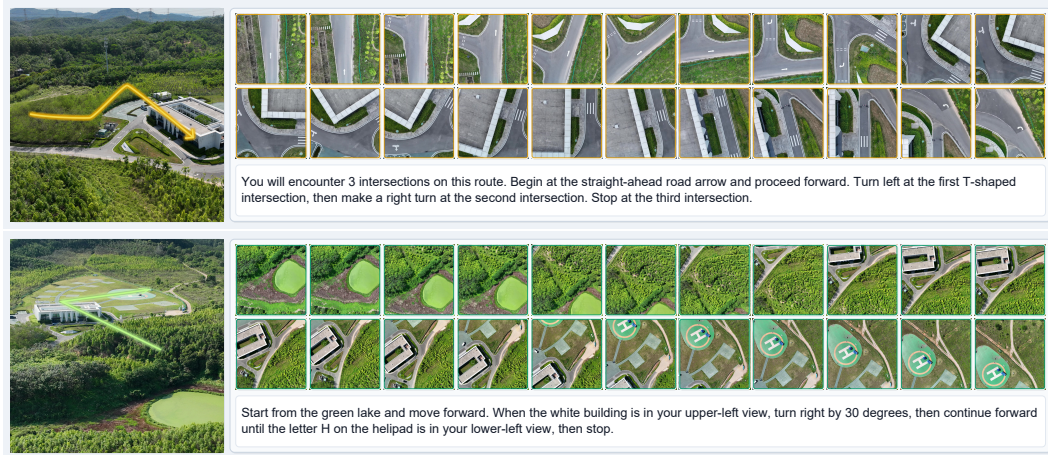

  \centering
  \begin{subfigure}{\textwidth}
    \centering
    \includegraphics[width=\textwidth,page=1]{appendices/figures/real-world/paper_figure_road_manual.pdf}
  \end{subfigure}
  \hfill
  \begin{subfigure}{\textwidth}
    \centering
    \includegraphics[width=\textwidth,page=1]{appendices/figures/real-world/paper_figure_landmark.pdf}
  \end{subfigure}
  \caption{Visualizations of real-world UAV navigation examples, including a \textbf{Route} task and a \textbf{Landmark} task.}
  \label{fig:traj_true}
\end{figure}

We conduct additional real-world flight tests to further examine the behavior of the satellite-trained model under physical deployment.
These tests cover all three SatNav task families: \textbf{Boundary}, \textbf{Landmark}, and \textbf{Route}.
Figure~\ref{fig:traj_true} presents examples for the \textbf{Landmark} and \textbf{Route} tasks, showing that the model can execute landmark-guided turning and route-following in real-world scenes.

\begin{figure}[p]
  \centering
  \begin{subfigure}{\textwidth}
    \centering
    \includegraphics[width=\textwidth,page=1]{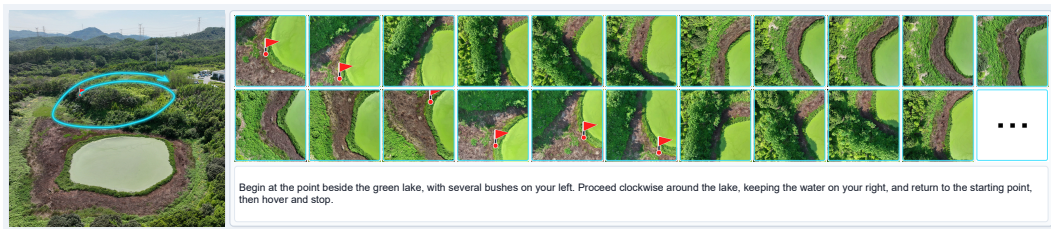}
  \end{subfigure}
  \caption{Representative real-world failure case on a \textbf{Boundary} task. The UAV is instructed to follow the boundary of the lake and stop after one complete loop, but fails to issue \texttt{stop} near the goal and continues circling for about two and a half loops.}
  \label{fig:traj_failed}
\end{figure}

Real-world deployment also exposes several challenging cases.
Some failures are caused by system-level factors, such as unstable network connections between the UAV, controller, and local workstation.
We also observe navigation challenges and failures that mirror those seen in satellite-map evaluation.
In particular, the model can approach or revisit the target region but fail to output \texttt{stop}.
Figure~\ref{fig:traj_failed} shows such a case: the UAV is instructed to follow the boundary around a lake and stop after completing the loop, but it continues circling instead of terminating autonomously.
These observations suggest that real-world deployment is feasible, while robust communication, stop calibration, and error recovery remain important directions for future work.

\section{Broader Impacts and Responsible Use}
\label{app:broader_impacts}
SatNav supports research on long-horizon UAV navigation for inspection, logistics, and environmental monitoring. Its scalable evaluation setting enables systematic analysis of visual grounding, memory, route following, and stopping behavior.

\paragraph{Intended use and risks.}
SatNav is intended for research and benchmark evaluation. UAV navigation capabilities can also be used for intrusive surveillance or military operations, raising privacy and physical safety concerns. Researchers should minimize the collection and retention of identifiable imagery and respect access restrictions around sensitive sites.

\paragraph{Safeguards for real-flight research.}
Real-flight studies should obtain applicable regulatory and site approvals and use controlled test areas with safe separation from uninvolved people and sensitive infrastructure. A qualified operator should supervise flights, with geofencing, immediate manual override, and predefined emergency procedures for communication loss, navigation errors, and low battery. Preflight checks should verify these safeguards and establish trial-abort conditions. Physical deployment also requires obstacle avoidance, altitude control, and safe recovery alongside the navigation capabilities evaluated by SatNav.

\section{Additional Experimental Results}
\label{app:extra_exps}

\subsection{Task-specific Baseline Performance}
\label{app:task_specific_baseline}

\begin{table*}[t]
\centering
\caption{Per-task diagnostic results on Test Seen and Test Unseen. SR and OS are reported as percentages. Best results are in \textbf{bold}, and second-best results are \underline{underlined}.}
\label{tab:per_task_diagnostic_results}
\setlength{\tabcolsep}{3.0pt}
\scriptsize

\makebox[\textwidth][c]{\small\bfseries (a) Test Seen}
\vspace{0.5pt}

\resizebox{\textwidth}{!}{%
\begin{tabular}{@{}l*{12}{c}@{}}
\toprule
\multirow{2}{*}{Model}
& \multicolumn{4}{c}{Boundary}
& \multicolumn{4}{c}{Landmark}
& \multicolumn{4}{c}{Route} \\
\cmidrule(lr){2-5} \cmidrule(lr){6-9} \cmidrule(lr){10-13}
& SR$\uparrow$ & OS$\uparrow$ & NE$\downarrow$ & Steps
& SR$\uparrow$ & OS$\uparrow$ & NE$\downarrow$ & Steps
& SR$\uparrow$ & OS$\uparrow$ & NE$\downarrow$ & Steps \\
\midrule
Seq2Seq
& 2.1 & 78.7 & 65.11 & 45.88
& 0.0 & 0.4 & 238.47 & 55.03
& 4.4 & 12.2 & 243.57 & 59.42 \\

CMA
& 12.4 & \textbf{92.4} & 63.54 & 80.05
& 0.0 & 2.9 & 789.61 & 159.76
& 13.9 & 32.0 & 214.13 & 59.58 \\

OpenFly
& 16.4 & 66.9 & 52.03 & 74.10
& 2.7 & 7.1 & 288.36 & 54.06
& 20.2 & 26.9 & 160.36 & 49.15 \\

OpenFly$^\ast$
& 27.0 & 71.5 & 41.47 & 69.20
& 6.6 & 9.9 & 311.11 & 59.18
& 29.3 & 33.2 & 137.69 & 45.82 \\

NaVILA
& 20.8 & 46.4 & 42.85 & 62.00
& 8.1 & 9.3 & 128.96 & 31.67
& 25.2 & 27.1 & 106.80 & 50.51 \\

NaVILA$^\ast$
& 40.5 & 67.3 & 35.85 & 74.23
& 4.5 & 5.3 & 146.61 & 30.85
& 29.8 & 32.5 & 98.90 & 49.44 \\

UniNaVid
& 13.0 & 76.3 & 59.90 & 102.00
& 22.2 & 48.4 & 377.89 & 71.01
& 39.7 & 56.7 & 88.85 & 47.52 \\

UniNaVid$^\ast$
& 36.2 & 75.2 & 27.92 & 63.68
& 58.9 & 68.8 & 186.74 & 53.37
& 53.8 & 60.7 & 47.85 & 45.22 \\

StreamVLN
& \textbf{68.9} & \underline{83.1} & \underline{15.38} & 63.77
& 60.0 & 64.7 & 98.10 & 46.38
& \underline{64.0} & 67.5 & 46.38 & 44.84 \\

StreamVLN$^\ast$
& \underline{68.5} & 79.3 & \textbf{14.88} & 63.18
& \underline{67.1} & \textbf{75.3} & \underline{92.24} & 46.24
& \textbf{66.3} & \textbf{69.2} & \textbf{35.60} & 47.09 \\

\midrule
\textbf{SwiftVLN}
& 62.8 & 75.4 & 15.71 & 58.97
& \textbf{71.6} & \underline{73.3} & \textbf{50.04} & 43.72
& 63.3 & \underline{68.5} & \underline{36.35} & 47.34 \\
\bottomrule
\end{tabular}%
}

\vspace{0.8em}

\makebox[\textwidth][c]{\small\bfseries (b) Test Unseen}
\vspace{0.5pt}

\resizebox{\textwidth}{!}{%
\begin{tabular}{@{}l*{12}{c}@{}}
\toprule
\multirow{2}{*}{Model}
& \multicolumn{4}{c}{Boundary}
& \multicolumn{4}{c}{Landmark}
& \multicolumn{4}{c}{Route} \\
\cmidrule(lr){2-5} \cmidrule(lr){6-9} \cmidrule(lr){10-13}
& SR$\uparrow$ & OS$\uparrow$ & NE$\downarrow$ & Steps
& SR$\uparrow$ & OS$\uparrow$ & NE$\downarrow$ & Steps
& SR$\uparrow$ & OS$\uparrow$ & NE$\downarrow$ & Steps \\
\midrule
Seq2Seq
& 2.0 & 72.8 & 76.08 & 49.30
& 0.2 & 0.3 & 276.37 & 61.46
& 2.5 & 14.5 & 308.90 & 64.97 \\

CMA
& 12.5 & \textbf{90.8} & 60.27 & 79.48
& 0.0 & 4.2 & 853.06 & 165.38
& 8.0 & 28.4 & 273.58 & 65.61 \\

OpenFly
& 12.1 & 60.4 & 61.47 & 78.78
& 5.1 & 12.9 & 289.76 & 61.49
& 17.6 & 23.7 & 234.09 & 59.20 \\

OpenFly$^\ast$
& 20.5 & 60.7 & 48.59 & 74.69
& 8.8 & 16.0 & 323.96 & 66.38
& 21.9 & 28.4 & 216.65 & 56.14 \\

NaVILA
& 14.6 & 40.8 & 57.22 & 62.21
& 3.5 & 4.6 & 149.17 & 32.95
& 20.6 & 25.5 & 177.50 & 65.40 \\

NaVILA$^\ast$
& 26.8 & 58.6 & 47.74 & 83.03
& 4.9 & 7.4 & 151.93 & 33.96
& 23.7 & 29.1 & 168.23 & 61.68 \\

UniNaVid
& 14.9 & 71.3 & 65.22 & 95.84
& 16.9 & 35.5 & 469.93 & 89.87
& 28.8 & 43.5 & 153.61 & 57.19 \\

UniNaVid$^\ast$
& 27.6 & 65.7 & 39.10 & 66.45
& 45.5 & 55.1 & 296.14 & 72.80
& 37.1 & 47.3 & 115.72 & 55.49 \\

StreamVLN
& \textbf{63.2} & \underline{79.1} & \underline{18.39} & 67.56
& 43.7 & 48.1 & \underline{145.66} & 54.06
& \underline{50.1} & 56.1 & 90.43 & 52.11 \\

StreamVLN$^\ast$
& \underline{62.9} & 77.7 & 19.68 & 66.74
& \textbf{59.0} & \textbf{65.0} & 152.97 & 58.98
& \textbf{53.7} & \textbf{62.6} & \textbf{87.02} & 58.04 \\

\midrule
\textbf{SwiftVLN}
& 58.0 & 73.0 & \textbf{17.56} & 63.09
& \underline{58.5} & \underline{63.1} & \textbf{78.01} & 50.84
& 45.1 & \underline{56.7} & \underline{90.03} & 58.18 \\
\bottomrule
\end{tabular}%
}

\vspace{2pt}
\begin{minipage}{\textwidth}
\footnotesize
\emph{Note.} $^\ast$ indicates models fine-tuned on SatNav from released navigation checkpoints; unmarked models are trained from their corresponding base backbones.
\end{minipage}
\end{table*}

Table~\ref{tab:per_task_diagnostic_results} further breaks down the navigation results by task, revealing distinct failure modes across Boundary, Landmark, and Route. The OS--SR gap measures the percentage of episodes that reach the success region but fail to terminate successfully there. On \textbf{Boundary}, the main challenge is not simply returning to the target region, but stopping at the correct time. For example, on the seen split, CMA achieves very high OS (92.4\%) but extremely low SR (12.4\%), together with long trajectories (80.05 steps), indicating that the agent often revisits the goal region but keeps moving, resulting in loop-like behaviors around the target instead of issuing \texttt{stop}. UniNaVid shows a similar tendency, with relatively long trajectories and a large OS-SR gap, suggesting inefficient termination even when the target region is reached. Seq2Seq exhibits a different termination failure: its OS is also much higher than SR (78.7\% vs. 2.1\%), but with substantially fewer steps (45.88), suggesting that it may pass through the target region yet stop later at an incorrect location rather than continuously circling. StreamVLN and SwiftVLN retain OS--SR gaps of 14.2 and 12.6 percentage points on the seen split, respectively, highlighting persistent stopping errors.

On \textbf{Landmark}, the dominant failure mode shifts from stopping to landmark localization. Classical baselines such as Seq2Seq and CMA nearly collapse on Landmark (achieving 0.0\% SR with NE exceeding 230 and 780, respectively), while the LVLM-based OpenFly-Agent also remains weak on this task. Their low SR and large NE suggest that these models often fail to ground the landmark and subsequently drift far away from the target. For other LVLM-based models, the OS-SR gap is much smaller than on Boundary (e.g., 4.7 percentage points for StreamVLN), suggesting that once the agent reaches the landmark region, it can usually stop; the harder part is reaching the correct landmark in the first place. The clear seen-to-unseen degradation (e.g., SwiftVLN's SR dropping from 71.6\% to 58.5\%) further shows that landmark localization remains sensitive to cross-city visual and spatial variations.

On \textbf{Route}, the main difficulty is long-horizon route following and intersection counting. Unlike Boundary, the OS-SR gap is relatively small for strong models, so failures are less about stopping after reaching the goal and more about whether the agent follows the correct route sequence. This is reflected by the sharp NE increase on unseen splits. For instance, NE increases from 35.60 to 87.02\,m for StreamVLN*, from 36.35 to 90.03\,m for SwiftVLN, and from 47.85 to 115.72\,m for UniNaVid*. These large increases suggest that intersection errors and counting mistakes accumulate over time: once the agent chooses a wrong branch or miscounts a turn, the final position can drift far from the target. Therefore, Route exposes limitations in maintaining sequential route state, especially under unseen route layouts.

Overall, the three tasks stress different navigation abilities and memory requirements. Boundary emphasizes retaining goal-arrival evidence and calibrating \texttt{stop}; Landmark requires reliable visual grounding and localization from long-term spatial cues; and Route depends on sequential route memory for long-horizon following and intersection counting.

\subsection{Effect of Base LVLM Backbones}
\label{app:base_backbone_effect}

\begin{table*}[t]
\centering
\caption{Performance comparison of base vision-language backbones on SatNav. SR, SPL, and OS are reported as percentages. Best results are in \textbf{bold}, and second-best results are \underline{underlined}.}
\label{tab:base_model_performance}
\setlength{\tabcolsep}{4.2pt}
\footnotesize
\resizebox{0.85\textwidth}{!}{%
\begin{tabular}{@{}llcccccccccc@{}}
\toprule
\multirow{2}{*}{Base Model}
& \multirow{2}{*}{Size}
& \multicolumn{5}{c}{Seen}
& \multicolumn{5}{c}{Unseen} \\
\cmidrule(lr){3-7} \cmidrule(lr){8-12}
& & SR$\uparrow$ & SPL$\uparrow$ & OS$\uparrow$ & NE$\downarrow$ & Steps
& SR$\uparrow$ & SPL$\uparrow$ & OS$\uparrow$ & NE$\downarrow$ & Steps \\
\midrule
\rowcolor{gray!10}
Qwen2.5-VL & 3B & 65.8 & 65.5 & 72.4 & \underline{34.05} & 49.97 & 53.7 & 53.2 & 64.1 & \underline{62.29} & 57.38 \\
Qwen2.5-VL & 7B & \textbf{70.2} & \textbf{69.8} & \underline{77.4} & \textbf{29.97} & 48.18 & \underline{56.6} & \underline{56.2} & \underline{66.2} & \textbf{55.33} & 54.62 \\
Qwen3-VL   & 2B & 67.5 & 66.7 & \textbf{79.5} & 52.64 & 58.13 & \textbf{57.0} & \textbf{56.3} & \textbf{70.3} & 71.94 & 65.84 \\
Qwen3-VL   & 8B & \underline{68.3} & \underline{67.6} & 76.7 & 45.12 & 52.16 & 56.0 & 55.3 & 64.9 & 69.79 & 58.55 \\
\bottomrule
\end{tabular}%
}
\vspace{-8pt}
\end{table*}

As mentioned in Section~\ref{sec:swiftvln_framework_overview}, SwiftVLN is designed as a flexible framework that can adapt different base LVLM backbones to the SatNav setting.
In the main experiments, the SwiftVLN reference model uses Qwen2.5-VL-3B as its backbone.
To verify the backbone extensibility of the framework, we replace the base model with LVLMs of different sizes and model families, including Qwen2.5-VL-7B and Qwen3-VL models.
All models are trained on the SatNav training split for one epoch using the same task interface, memory configuration, action space, and training protocol as the SwiftVLN reference model, and are then evaluated on the same SatNav splits.
Table~\ref{tab:base_model_performance} reports the resulting performance.
This experiment is intended to demonstrate that SwiftVLN supports consistent training and evaluation across different LVLM backbones.

\subsection{Transfer Evaluation with Rendered UAV Observations}
\label{app:rendered_transfer}

\begin{table*}[t]
\centering
\caption{Navigation performance on 94 episodes rendered from real-world 3D assets. SR, SPL, and OS are reported as percentages, and NE in meters.}
\label{tab:rendered_transfer}
\small
\setlength{\tabcolsep}{5pt}
\renewcommand{\arraystretch}{1.15}
\begin{tabular}{llrrrr}
\toprule
Method & Image perturbation & SR$\uparrow$ & SPL$\uparrow$ & OS$\uparrow$ & NE$\downarrow$ \\
\midrule
SwiftVLN & None & 34.0 & 33.5 & 47.9 & 103.26 \\
SwiftVLN w/ adapter & None & \textbf{43.6} & \textbf{42.3} & \textbf{59.6} & \textbf{62.33} \\
\midrule
SwiftVLN & Exposure + Jitter & 28.7 & 27.8 & 44.7 & 121.02 \\
SwiftVLN w/ adapter & Exposure + Jitter & \textbf{39.4} & \textbf{38.1} & \textbf{56.4} & \textbf{70.64} \\
\bottomrule
\end{tabular}
\end{table*}

Aside from real-world experiments in Section~\ref{sec:real_world_demo}, we further evaluate satellite-to-UAV transfer using nadir observations rendered from real-world 3D assets. Four low-altitude UAV experts design 94 navigation episodes using the Birmingham and Cambridge point clouds from SensatUrban~\citep{hu2022sensaturban} and the Office, Road, and City 3D Gaussian Splatting (3DGS) assets from HUGE-Bench~\citep{guo2026huge}. The instructions are processed using the same LLM-based rewriting pipeline as SatNav. 

We implement two rendering backends, \texttt{pcdsim} and \texttt{3dgssim}, to generate nadir UAV observations from point clouds and 3DGS scenes, respectively. Both backends use perspective cameras and 3D scene geometry. We evaluate the satellite-trained SwiftVLN reference policy with and without the visual adapter described in Appendix~\ref{app:swiftvln_sim2real_details} on the 94 episodes.
We further consider two observation conditions: clean rendered views and views with exposure and jitter perturbations. This comparison examines how visual adaptation affects navigation performance under changes in observation.

Table~\ref{tab:rendered_transfer} shows that the adapter improves navigation performance under both clean and perturbed observations, supporting its effectiveness for satellite-to-UAV transfer. Exposure and jitter reduce performance for both variants, highlighting the limitations of the current adaptation approach and the substantial room for improvement in satellite-to-UAV transfer.

\clearpage
\begin{table*}[!t]
\centering
\caption{Sensitivity of SwiftVLN to benchmark configuration. Each variant changes one factor from the default setting. Best SR results are in \textbf{bold}, and second-best results are \underline{underlined}.}
\label{tab:configuration_sensitivity}
\small
\setlength{\tabcolsep}{4pt}
\renewcommand{\arraystretch}{1.15}
\begin{tabular}{llrrrr}
\toprule
Factor & Setting & Train time & Eval time & Seen SR & Unseen SR \\
\midrule
Baseline & Default & 4:19:45 & 2:24:12 & \underline{65.8} & \underline{53.7} \\
\midrule
Satellite resolution & Zoom Level 17 & 4:24:19 & 2:29:25 & 64.9 & 52.8 \\
 & Zoom Level 15 & 4:12:06 & 2:23:27 & 62.4 & 52.5 \\
\midrule
Crop size & $336\times336$ px & 2:37:35 & 2:37:06 & 65.0 & 52.0 \\
 & $224\times224$ px & 1:33:34 & 2:50:05 & 51.1 & 40.8 \\
\midrule
Spatial coverage & $50\times50$\,m & 4:16:34 & 3:02:32 & 56.4 & 46.3 \\
 & $200\times200$\,m & 3:56:40 & 3:42:08 & 56.9 & 50.7 \\
\midrule
Action granularity & 5\,m / $7.5^\circ$ & 8:01:56 & 4:54:48 & \textbf{66.7} & \textbf{56.9} \\
\bottomrule
\end{tabular}
\end{table*}

\subsection{Sensitivity to Benchmark Configuration}
\label{app:configuration_sensitivity}

We examine how satellite-image resolution, crop size, spatial coverage, and action granularity affect navigation performance and computational cost. Starting from the default SatNav configuration, we vary one factor at a time and retrain and evaluate SwiftVLN using eight NVIDIA H100 GPUs. The default setting uses satellite tiles at zoom level 19, $448\times448$-pixel observations covering $100\times100$\,m, a forward step of 10\,m, and a turn angle of $15^\circ$.

Spatial coverage is varied through the ground footprint of the satellite crop. The $50\times50$\,m and $200\times200$\,m footprints correspond to nominal altitudes of 25\,m and 100\,m, respectively, under the $90^\circ$ horizontal field of view used in our observation model. Crop-size variants change the observation dimensions in pixels while retaining the default ground footprint.

Table~\ref{tab:configuration_sensitivity} shows that reducing the crop size to $224\times224$ pixels or changing spatial coverage substantially lowers SR, while lower satellite-image resolution causes smaller decreases. Finer actions improve SR on both splits but roughly double training and evaluation time, illustrating the trade-off between navigation accuracy and computational cost.

\clearpage
\begin{figure}[p]
  \centering
  \begin{subfigure}{\textwidth}
    \centering
    \includegraphics[width=\textwidth,page=1]{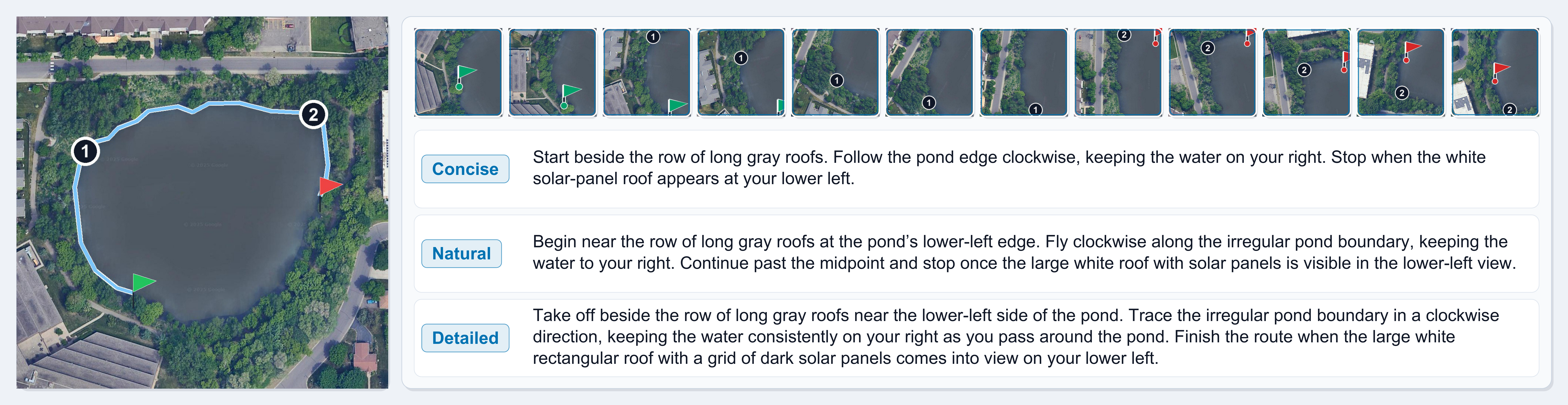}
  \end{subfigure}
  \hfill
  \begin{subfigure}{\textwidth}
    \centering
    \includegraphics[width=\textwidth,page=1]{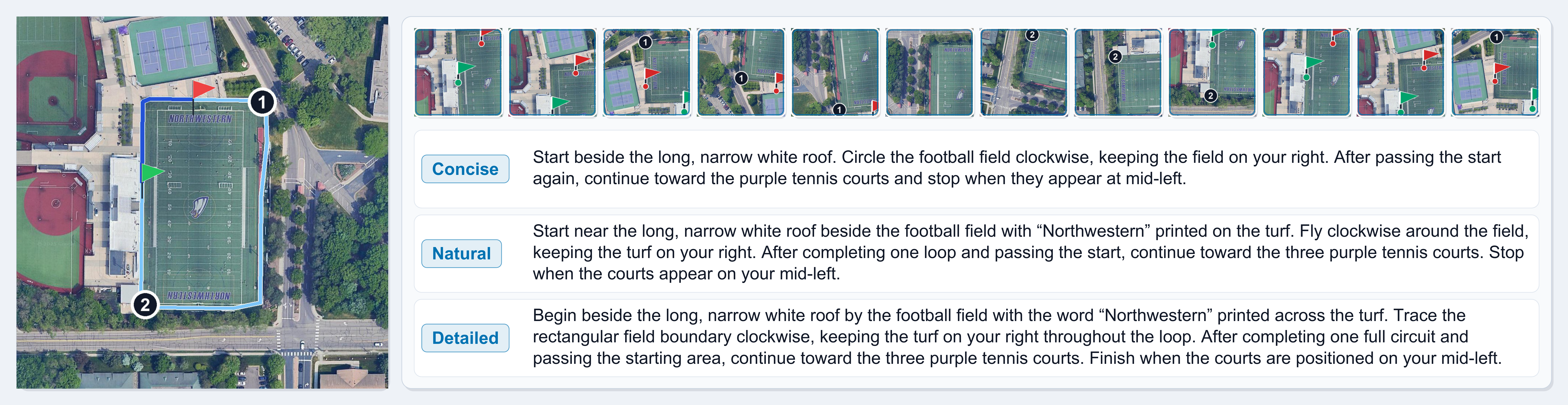}
  \end{subfigure}
  \hfill
  \begin{subfigure}{\textwidth}
    \centering
    \includegraphics[width=\textwidth,page=1]{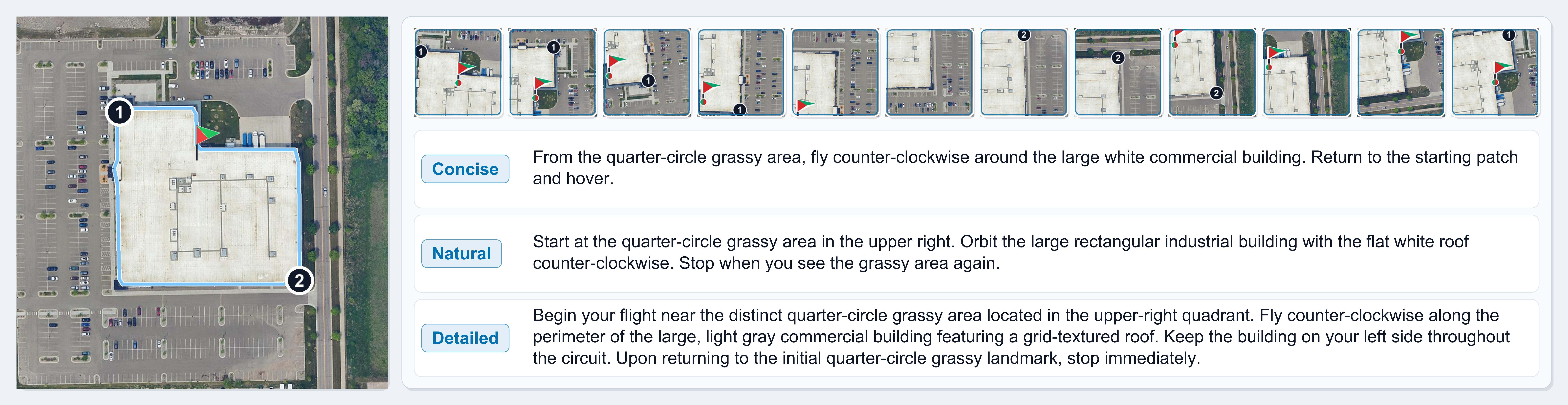}
  \end{subfigure}
  \caption{Visualizations of trajectory examples for the Boundary task family.}
  \label{fig:traj_boundary}
\end{figure}

\begin{figure}[p]
  \centering
  \begin{subfigure}{\textwidth}
    \centering
    \includegraphics[width=\textwidth,page=1]{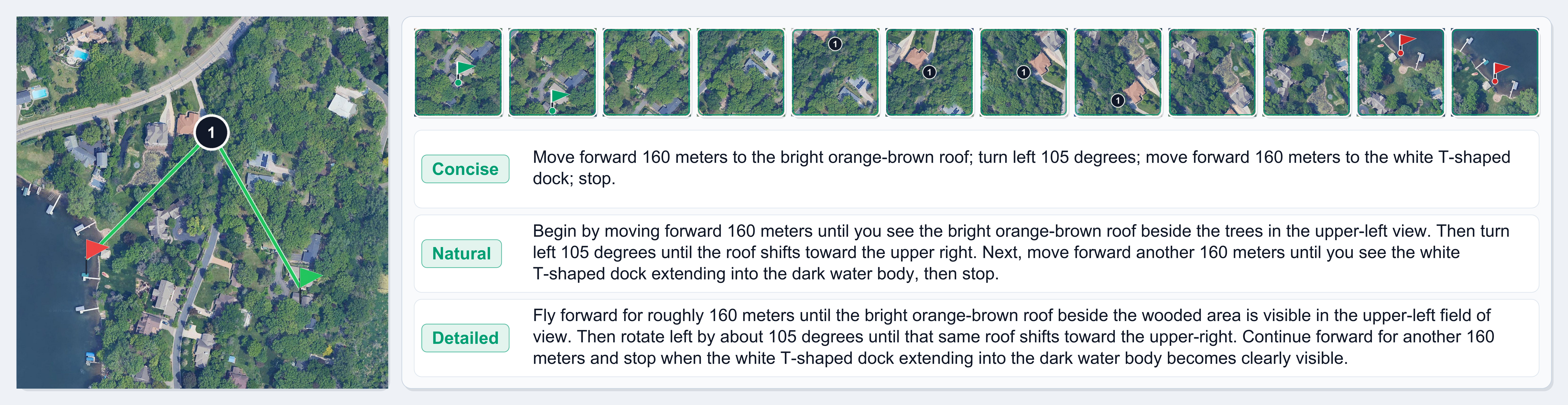}
  \end{subfigure}
  \hfill
  \begin{subfigure}{\textwidth}
    \centering
    \includegraphics[width=\textwidth,page=1]{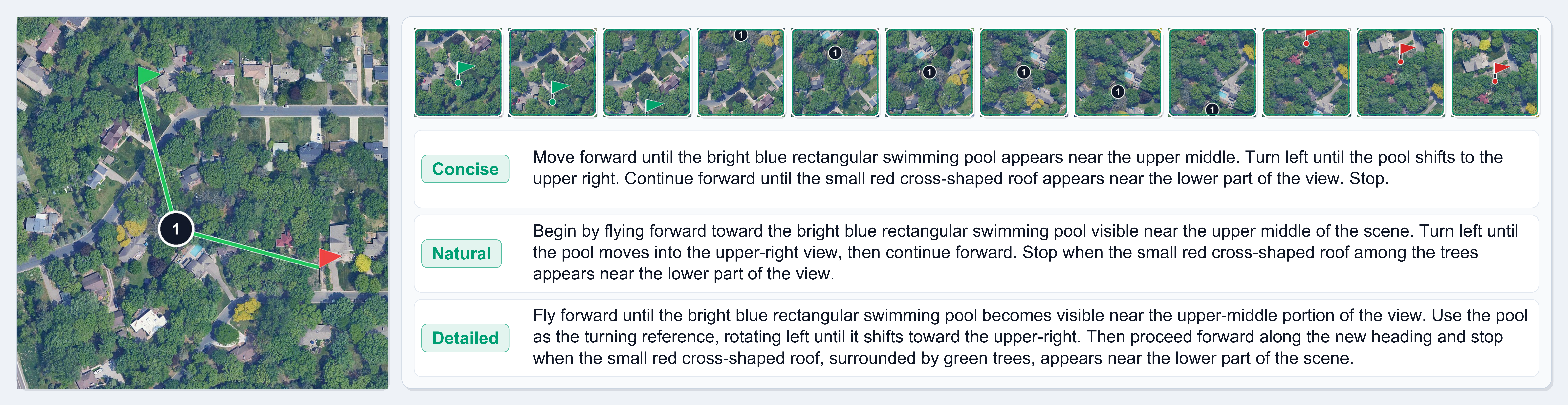}
  \end{subfigure}
  \caption{Visualizations of trajectory examples for the Landmark task family.}
  \label{fig:traj_landmark}
\end{figure}

\begin{figure}[p]
  \centering
  \begin{subfigure}{\textwidth}
    \centering
    \includegraphics[width=\textwidth,page=1]{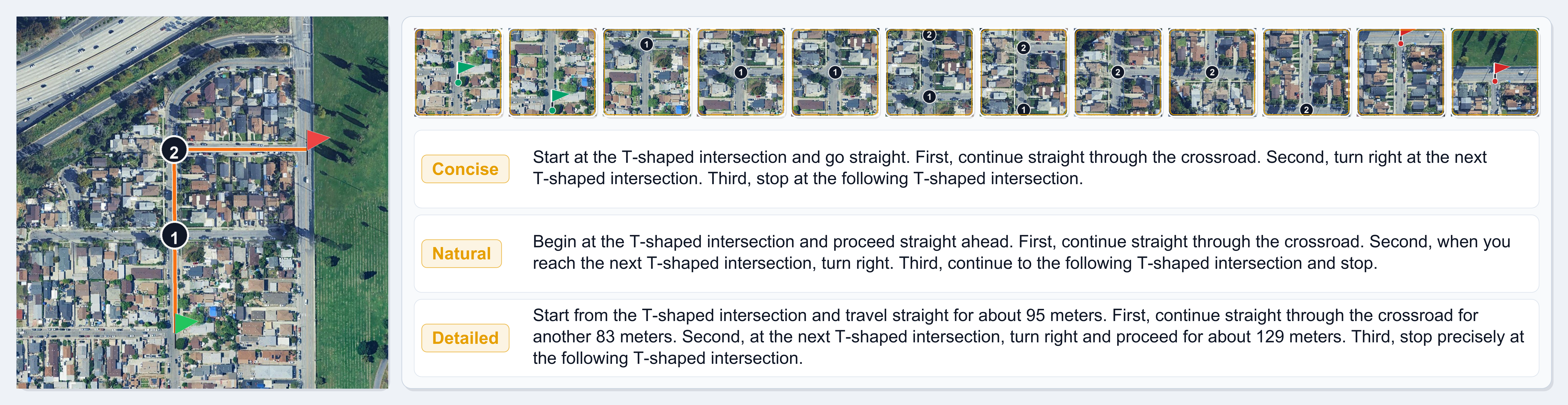}
  \end{subfigure}
  \hfill
  \begin{subfigure}{\textwidth}
    \centering
    \includegraphics[width=\textwidth,page=1]{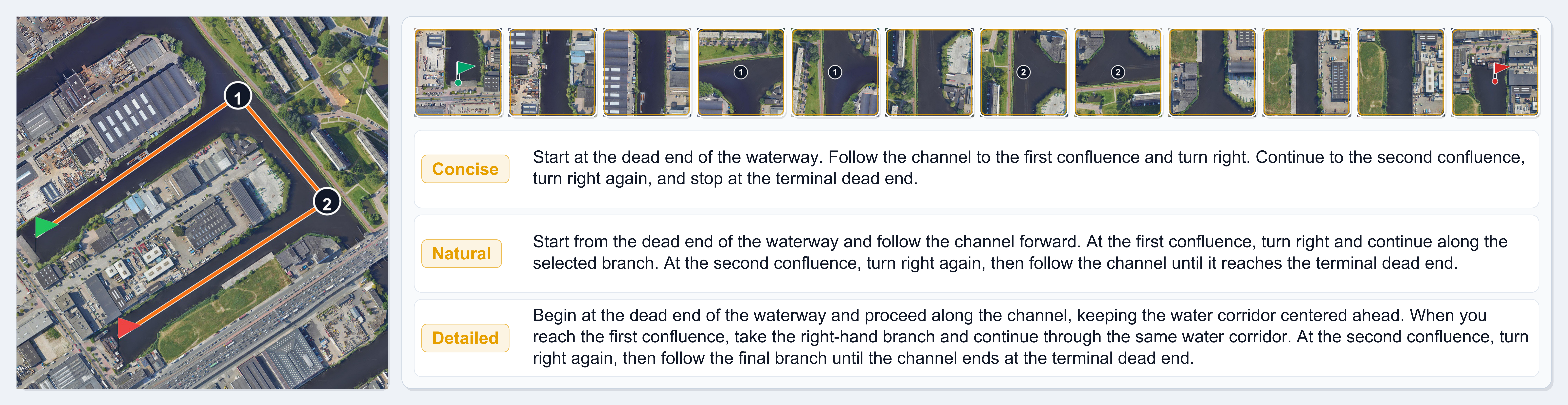}
  \end{subfigure}
  \caption{Visualizations of trajectory examples for the Route task family.}
  \label{fig:traj_road}
\end{figure}

\end{document}